\documentclass[journal]{IEEEtran}
\usepackage{multirow}
\usepackage{amssymb,amsmath}
\usepackage{rotating}
\usepackage[ruled,vlined]{algorithm2e}
\usepackage{algorithmic}
\usepackage{bm}
\usepackage{graphicx}
\usepackage{subfigure}
\usepackage{color}
\usepackage{amsthm}
\usepackage{amsbsy}

\usepackage{tabularx}
\usepackage{multirow}

\ifCLASSINFOpdf
\else
\fi
\begin{document}
%
\title{Residual Networks of Residual Networks:\\ Multilevel Residual Networks}
%
%
%

\author{Ke~Zhang,~\IEEEmembership{Member,~IEEE,}
        Miao~Sun,~\IEEEmembership{Student Member,~IEEE,}
        Tony~X.~Han,~\IEEEmembership{Member,~IEEE,}
        Xingfang~Yuan,~\IEEEmembership{Student Member,~IEEE,}
        Liru~Guo,
        and~Tao~Liu
\thanks{This work is supported by National Natural Science Foundation of China (Grants No. 61302163, No. 61302105 and No. 61501185), Hebei Province Natural Science Foundation (Grants No. F2015502062 and No. F2016502062) and the Fundamental Research Funds for the Central Universities (Grants No. 2016MS99).}
\thanks{K. Zhang is with the Department
of Electronic and Communication Engineering, North China Electric Power University, Baoding,
Hebei, 071000 China e-mail: zhangkeit@ncepu.edu.cn.}
\thanks{M. Sun is with the Department
	of Electrical and Computer Engineering, University of Missouri, Columiba,
	MO, 65211 USA e-mail: msqz6@mail.missouri.edu.}
\thanks{T. X. Han is with the Department
	of Electrical and Computer Engineering, University of Missouri, Columiba,
	MO, 65211 USA e-mail: HanTX@missouri.edu.}
\thanks{X. Yuan is with the Department
	of Electrical and Computer Engineering, University of Missouri, Columiba,
	MO, 65211 USA e-mail: xyuan@mail.missouri.edu.}
\thanks{L. Guo is with the Department
	of Electronic and Communication Engineering, North China Electric Power University, Baoding,
	Hebei, 071000 China e-mail: glr9292@126.com.}
\thanks{T. Liu is with the Department
	of Electronic and Communication Engineering, North China Electric Power University, Baoding,
	Hebei, 071000 China e-mail: taoliu@ncepu.edu.cn.}
\thanks{Manuscript received 15 Aug.2016; revised  20 Dec.2016.}}

%
%


\markboth{IEEE Transactions on \LaTeX\ Class Files,~Vol.~14, No.~8, August~2016}%
{Shell \MakeLowercase{\textit{et al.}}: Bare Demo of IEEEtran.cls for IEEE Journals}
%



\maketitle

\begin{abstract}
A residual-networks family with hundreds or even thousands of layers dominates major image recognition tasks, but building a network by simply stacking residual blocks inevitably limits its optimization ability. This paper proposes a novel residual-network architecture, Residual networks of Residual networks (RoR), to dig the optimization ability of residual networks. RoR substitutes optimizing residual mapping of residual mapping for optimizing original residual mapping. In particular, RoR adds level-wise shortcut connections upon original residual networks to promote the learning capability of residual networks. More importantly, RoR can be applied to various kinds of residual networks (ResNets, Pre-ResNets and WRN) and significantly boost their performance. Our experiments demonstrate the effectiveness and versatility of RoR, where it achieves the best performance in all residual-network-like structures. Our RoR-3-WRN58-4+SD models achieve new state-of-the-art results on CIFAR-10, CIFAR-100 and SVHN, with test errors 3.77\%, 19.73\% and 1.59\%, respectively. RoR-3 models also achieve state-of-the-art results compared to ResNets on ImageNet data set.
\end{abstract}


%
\IEEEpeerreviewmaketitle

\section{Introduction}
%
%
%
%
\IEEEPARstart{C}{onvolutional} Neural Networks (CNNs) have given the computer vision community a significant shock~\cite{lecun2015deeplearning}, and have been improving state-of-the-art results in many computer vision applications. Since AlexNets'~\cite{Krizhenvshky2012dcn} ground-breaking victory at the ImageNet Large Scale Visual Recognition Challenge 2012 (ILSVRC 2012)~\cite{Russ2014imagenetchallenge}, deeper and deeper CNNs~\cite{Krizhenvshky2012dcn,miao2014detection,lin2013NiN,sermanet2013overfeat,simonyan2014vgg,romero2014fitnets,lee2015dsn,springgenberg2014allcnn,szegedy2015googlenet,he2015resnets} have been proposed and achieved better performance on ImageNet or other benchmark data sets. The results of these models revealed 
\begin{figure}
\centering
\includegraphics[width=1.0\linewidth]{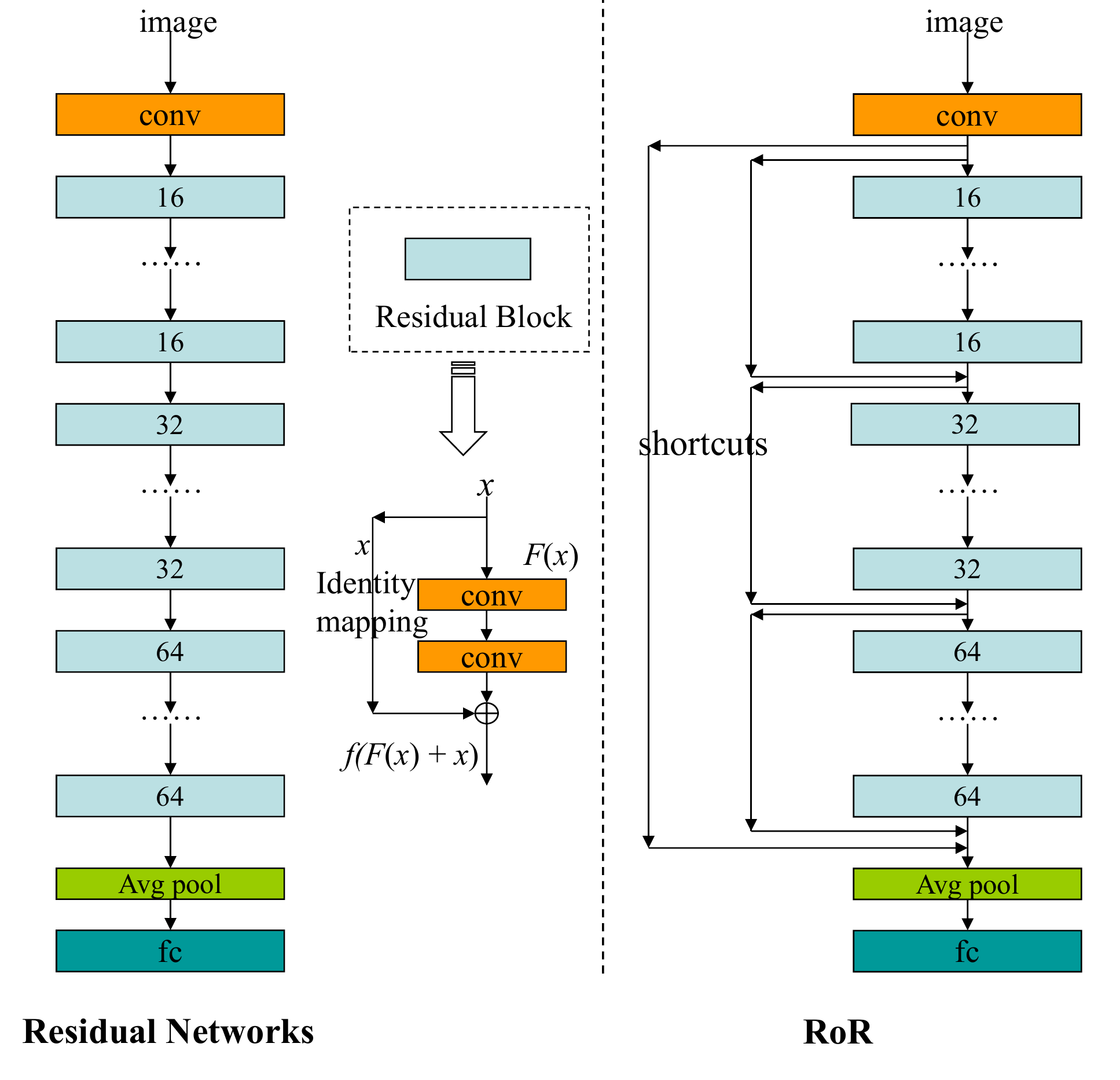}
\caption{The image on the left of the dashed line is an original residual network, which contains a series of residual blocks, and each residual block has one shortcut connection. The number (16, 32, or 64) on each residual block is the number of output feature map. $F(x)$ is the residual mapping and $x$ is the identity mapping. The original mapping is represented as $F(x)+x$. The image on the right of the dashed line is our new residual networks of residual networks architecture with three shortcut levels. RoR is constructed by adding identity shortcuts level by level based on original residual networks.}
\label{fig:basicnetworks}
\end{figure}

the importance of network depth, as deeper networks lead to superior results.
\par
With a dramatic increase in depth, Residual Networks (ResNets)~\cite{he2015resnets} achieved the state-of-the-art performance award at the ILSVRC 2015 for classification, localization, detection, and COCO detection as well as segmentation tasks. However, very deep models will suffer vanishing gradients and overfitting problems; Thus, the performance of thousand-layer ResNets is worse than hundred-layer ResNets. Then the Identity Mapping ResNets (Pre-ResNets)~\cite{he2016preresnets} simplified the residual networks training by BN-ReLU-conv order. Pre-ResNets can alleviate the vanishing gradients problem, so that the performance of thousand-layer Pre-ResNets can be further improved. The latest Wide Residual Networks (WRN)~\cite{zagoruyko2016wrn} treated the vanishing gradients problem by decreasing depth and increasing the width of residual networks. Nevertheless, the exponentially increasing number of parameters brought by broader networks worsens the overfitting problem. As a result, dropout and drop-path methods are usually used to alleviate overfitting, and the leading method on ResNets is Stochastic Depth residual networks (SD)~\cite{huang2016SD}, which can improve test accuracy and reduce training time. All kinds of residual networks are based on one basic hypothesis: By using shortcut connections, residual networks perform residual mapping fitted by stacked nonlinear layers, which is easier to be optimized than the original mapping~\cite{he2015resnets}. Furthermore, we hypothesize that the residual mapping of residual mapping is easier to optimize than original residual mapping. So on top of this hypothesis, we can construct a better residual-network architecture to enhance performance.
\par
In this paper, we presented a novel and simple architecture called Residual networks of Residual networks (RoR). First, compared to the original one-level-shortcut-connection ResNets, we added extra shortcut connections to the original ResNets level by level. A multilevel network was then constructed, as seen in Fig.~\ref{fig:basicnetworks}. For this network, we analyzed the effects of different shortcut level numbers, shortcut types and maximum epoch numbers. Second, to alleviate overfitting, we trained RoR with the drop-path method, and obtained an apparent performance boost. Third, we built RoR on various residual networks (Pre-ResNets and WRN), and found that RoR is not only suitable for original ResNets, but also fits in nicely with other residual networks. Through massive experiments on CIFAR-10, CIFAR-100~\cite{krizhevsky2009cifar}, SVHN~\cite{netzer2011SVHN} and ImageNet~\cite{Russ2014imagenetchallenge}, our RoR obtained as much optimization ability of the residual networks as possible, only by adding a few shortcuts. Although this approach seems quite simple, it is surprisingly effective in practice and achieves the new state-of-the-art results on the above data sets. 
\par
Our main contribution is threefold:
\par
 1. Our introduction of RoR improved the optimization ability of ResNets by adding a few identity shortcuts. And RoR achieved better performance than ResNets by using the same number of layers on different data sets.
\par 
 2. RoR is suitable for other residual networks as well and will boost their performance, which makes it an effective complement of the residual-networks family.
 \par 
 3. Through experiments, we analyzed the effects of different depths, widths, shortcuts level numbers, shortcut types and maximum epoch numbers to RoR, and developed reasonable strategies for RoR applications in different data sets.
\par
 The remainder of the paper is organized as follows. Section II briefly reviews related work for deep convolutional neural networks and the residual-networks family. The proposed RoR method is illustrated in Section III. The optimization of RoR is described in Section IV. Experimental results and analysis are presented in Section V, leading to conclusions in Section VI. 

\section{Related Work}

\subsection{Deeper and Deeper Convolutional Neural Networks}


In the past several years, deeper and deeper CNNs have been constructed, as the more convolutional layers are in CNNs, the better optimization ability CNNs can obtain. From 5-conv+3-fc AlexNet (ILSVRC2012 winner)~\cite{Krizhenvshky2012dcn} to the 16-conv+3-fc VGG networks~\cite{simonyan2014vgg} and 21-conv+1-fc GoogleNet (ILSVRC2014 winner)~\cite{szegedy2015googlenet}, both the accuracy and the depth of CNNs have continued to increase. However, very deep CNNs face the crucial problem of vanishing gradients~\cite{bengio2014vanish}. Earlier works adopted initialization methods and layer-wise training to reduce this problem~\cite{glorot2010earlywork,erhan2010earlywork}. Moreover, the ReLU activation function~\cite{nair2010relu} and its variants ELU~\cite{clevert2015elu}, PReLU~\cite{he2015prelu}, and PELU~\cite{trottier2016pelu} also prevent vanishing gradients. Fortunately, this problem could be largely addressed by batch normalization (BN)~\cite{ioffe2015bn} and carefully normalized weights initialization~\cite{he2015prelu,mishkin2015initial} according to recent research. BN~\cite{ioffe2015bn} standardized the mean and variance of hidden layers for each mini-batch, while MSR~\cite{he2015prelu} initialized the weights by a more reasonable variance. On the other hand, a degradation problem has been exposed~\cite{he2015resnets,he2015timecost,sriva2015highway} that is, not all systems are easy to optimize. In order to resolve this problem, several methods were proposed. Highway Networks~\cite{sriva2015highway} consisted of a mechanism allowing 2D-CNNs to interact with a simple memory mechanism. Even with hundreds of layers, highway networks can be trained directly through simple gradient descent. ResNets~\cite{he2015resnets} simplified Highway Networks using a simple skip connection mechanism to propagate information to deeper layers of networks. ResNets are simpler and more effective than highway Networks. Recently, FractalNet~\cite{larsson2016fractalnet} generated an extremely deep network whose structural layout was precisely a truncated fractal by repeating application of a single expansion rule, and this method showed that residual learning was not required for ultra-deep networks. However, in order to get the competitive performance of ResNets, FractalNet must have many more parameters than ResNets. Hence, more and more residual network variants and architectures have been proposed, and they form a residual-networks family together~\cite{he2016preresnets,zagoruyko2016wrn,huang2016SD,trottier2016pelu,targ2016rir,shah2016elu,singh2016swapout,shen2016wresnets,moniz2016crmn}.
\subsection{Residual-Networks Family}
The basic idea of ResNets~\cite{he2015resnets} is that residual mapping is easy to optimize, so ResNets skip blocks of convolutional layers by using shortcut connections to form shortcut blocks (residual blocks). These stacked residual blocks greatly improve training efficiency and largely resolve the degradation problem by employing BN~\cite{ioffe2015bn} and MSR~\cite{he2015prelu}. The ResNets architecture and residual blocks are shown in Fig.~\ref{fig:basicnetworks}, where each residual block can be expressed in a general form:
\begin{equation}
\begin{array}{cl}
y_{l}=h(x_{l})+F(x_{l}, W_{l}),\\
x_{l+1}=f(y_{l}
)\end{array}
\label{E:ResNets1}
\end{equation}
where $x_{l}$ and $x_{l}+1$ are input and output of the $l$-th block, respectively. $F$ is a residual mapping function, $h(x_{l})=x_{l}$ is an identity mapping function, and $f$ is a ReLU function. However, the vanishing gradients problem still exists, as the test result of 1202-layer ResNets is worse than 110-layer ResNets on CIFAR-10~\cite{he2015resnets}. 
\par 
In the Pre-ResNets, He et al.~\cite{he2016preresnets} created a ``direct'' path for propagating information through the entire network by letting both $h(x_{l})$ and $f$ serve as identity mappings. The residual block of Pre-ResNets performs the follow computation:
\begin{equation}
\begin{array}{cl}
x_{l+1}=h(x_{l})+F(x_{l}, W_{l})
\end{array}
\label{E:ResNets2}
\end{equation}
The new residual block with a BN-ReLU-conv order can reduce training difficulties, so that Pre-ResNets can get better performance than original ResNets. More importantly, Pre-ResNets can reduce the vanishing gradients problem even for 1001-layer Pre-ResNets. Inspired by Pre-ResNets, Shen et al.~\cite{shen2016wresnets} proposed weighted residuals for very deep networks (WResNet), which removed the ReLU from highway and used weighted residual functions to create a ``direct'' path. This method is also capable of 1000+ layers residual networks training and achieves good accuracy. In order to further reduce vanishing gradients, Shah et al.~\cite{shah2016elu}, Trottier et al.~\cite{trottier2016pelu} proposed the use of ELU and PELU respectively instead of ReLU in residual networks. 
\par 
In addition to the vanishing gradients problem, overfitting is another challenging issue for CNNs. Huang and Sun et al.~\cite{huang2016SD} proposed a drop-path method, Stochastic Depth residual networks (SD), which randomly dropped a subset of layers and bypassed them with identity mapping for every mini-batch. SD can alleviate overfitting and reduce vanishing problem, so it is a good complement of residual networks. By combining dropout and SD, Singh et al.~\cite{singh2016swapout} proposed a new stochastic training method, SwapOut, which can be viewed as an ensemble of ResNets, dropout ResNets, and SD ResNets. 
\par
Recently, more variants of residual networks have been proposed, and they all promote learning capability by expending width of the model. Resnet in Resnet (RiR)~\cite{targ2016rir} is a generalized residual architecture which combines ResNets and standard CNNs in parallel residual and non-residual streams. WRN~\cite{zagoruyko2016wrn} decreases depth and increases width of residual networks by adding more feature planes, and achieves the latest state-of-the-art results on CIFAR-10 and SVHN. The newest Convolutional Residual Memory Networks (CRMN)~\cite{moniz2016crmn} was inspired by WRN and Highway Networks. CRMN augments convolutional residual networks with a long short-term memory mechanism based on WRN, and achieves the latest state-of-the-art performance on CIFAR-100. These wider residual networks indicates that wide and shallow residual networks result in good performance and easy training. 
\section{Residual Networks of Residual Networks}
RoR is based on a hypothesis: To dig the optimization ability of residual networks, we can optimize the residual mapping of residual mapping. So we add shortcuts level by level to construct RoR based on residual networks.
\par
\begin{figure}
\centering
\includegraphics[width=1.0\linewidth]{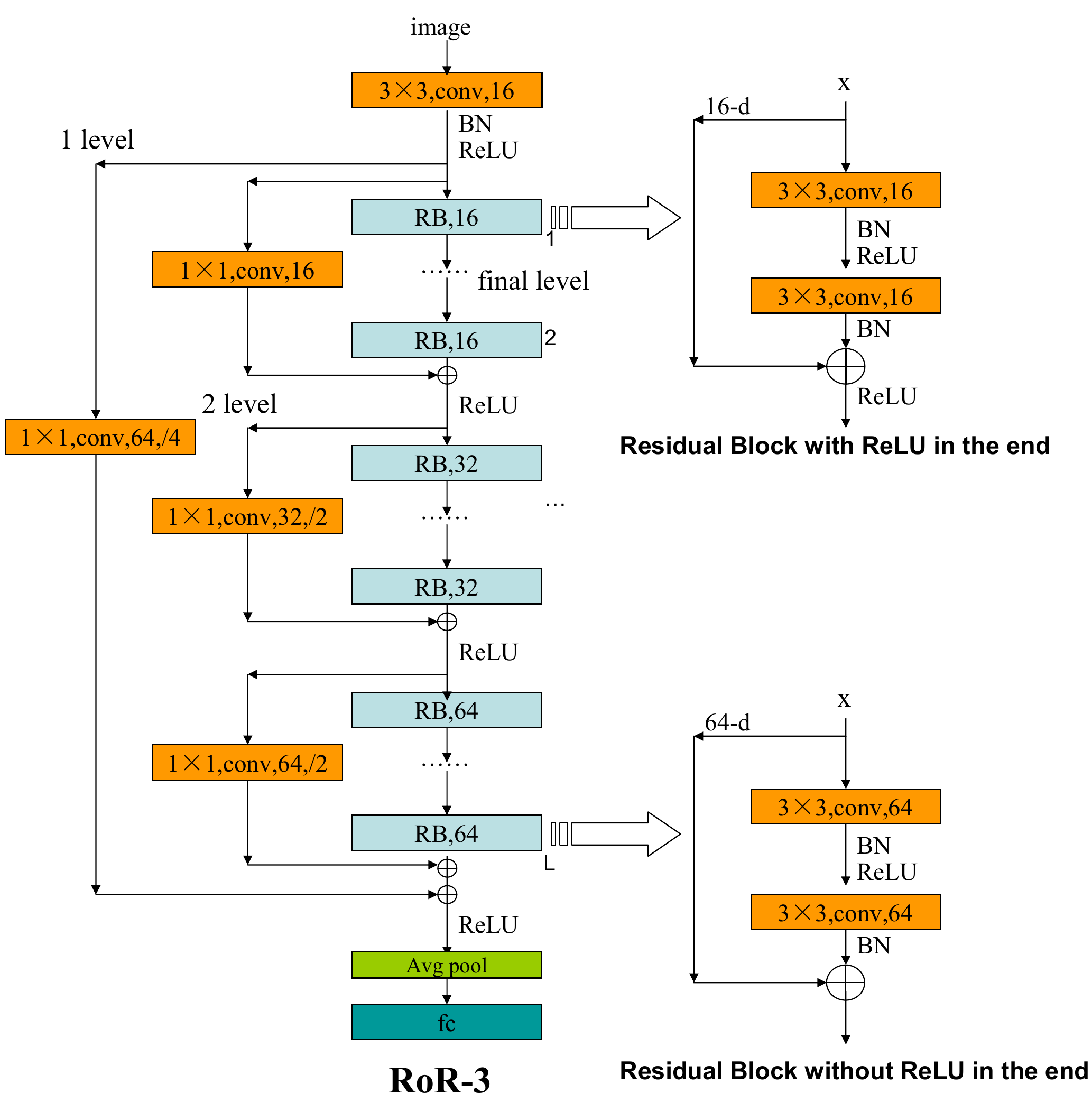}
\caption{RoR-3 architecture. The shortcut on the left is a root-level shortcut, and the remaining shortcuts are made up of three orange shortcuts, which are middle-level shortcuts. The blue shortcuts are final-level shortcuts. ReLU is followed by an addition. The projection shortcut is done by 1$\times$1 convolutions. Finally, RoR adopts a conv-BN-ReLU order in residual blocks.}
\label{fig:RoRnetworks}
\end{figure}

 
Fig.~\ref{fig:RoRnetworks} shows the original residual network with $L$ residual blocks. These $L$ original residual blocks are denoted as the final-level shortcuts. First, we add a shortcut above all residual blocks, and this shortcut can be called a root shortcut or first-level shortcut. Generally, we use 16, 32 and 64 filters sequentially in the convolutional layers~\cite{he2015resnets,he2016preresnets}, and each kind of filter has $L$/3 residual blocks which form three residual block groups. Second, we add a shortcut above each residual block group, and these three shortcuts are called second-level shortcuts. Then we can continue adding shortcuts as inner-level shortcuts by separating each residual block group equally. Finally, the shortcuts in the original residual blocks are regarded as the final-level shortcuts. Let $m$ denote a shortcut level number, $m$=1, 2, 3.... When $m$=1, RoR is an original residual networks with no other shortcut level. When $m$=2, the RoR has a root level and a final level. In this paper, $m$ is 3, so the RoR has a root level, middle level and final level, as shown in Fig.~\ref{fig:RoRnetworks}. Compared to the top-right residual block,the bottom-right block is without ReLU in the end, because there are extra additions following it.
\begin{figure}
\centering
\includegraphics[width=1.0\linewidth]{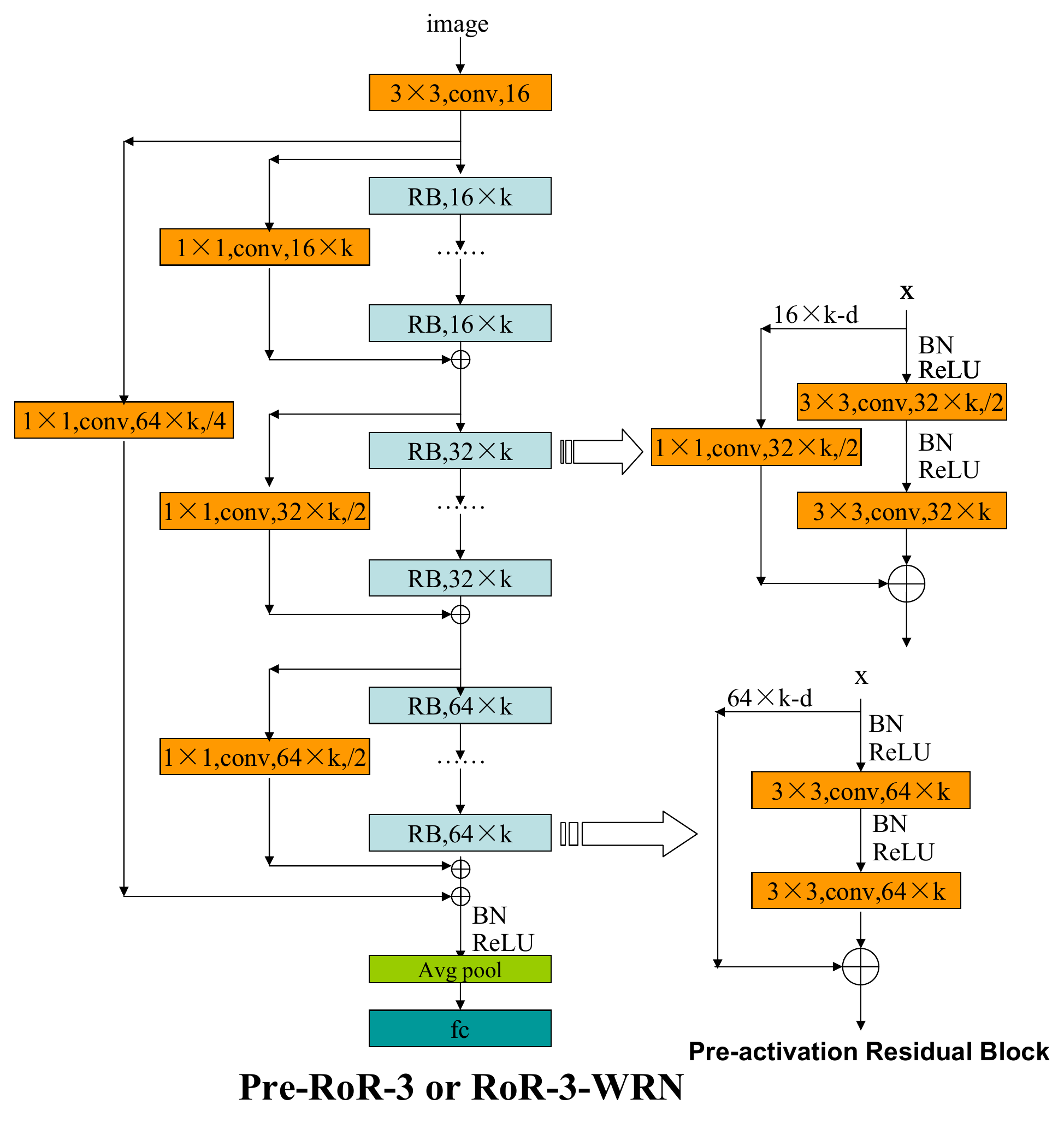}
\caption{Pre-RoR-3 and RoR-3-WRN architectures. $m$=3. The addition is followed by the ReLU. Projection shortcut is done by 1$\times$1 convolutions. BN-ReLU-conv order in residual blocks is adopted. If $k$=1, this is a Pre-RoR-3 architecture; Otherwise, this is an RoR-3-WRN architecture. There are several ``direct'' paths for propagating information created by identity mappings.}
\label{fig:Pre-RoRnetworks}
\end{figure}
\par
When $m$=3, three residual blocks located at the end of each residual block group can be expressed by the following formulations, and the other original residual blocks remain the same. 
\begin{equation}
\begin{array}{cl}
y_{L/3}=g(x_{1})+h(x_{L/3})+F(x_{L/3}, W_{L/3}),\\
x_{L/3+1}=f(y_{L/3}
)\end{array}
\label{E:RoRNets1}
\end{equation}
\begin{equation}
\begin{array}{cl}
y_{2L/3}=g(x_{L/3+1})+h(x_{2L/3})+F(x_{2L/3}, W_{2L/3}),\\
x_{2L/3+1}=f(y_{2L/3}
)\end{array}
\label{E:RoRNets2}
\end{equation}
\begin{equation}
\begin{array}{cl}
y_{L}=g(x_{1})+g(x_{2L/3+1})+h(x_{L})+F(x_{L}, W_{L}),\\
x_{L+1}=f(y_{L}
)\end{array}
\label{E:RoRNets3}
\end{equation}
where $x_{l}$ and $x_{l}$+1 are input and output of the $l$-th block, and $F$ is a residual mapping function; $h(x_{l})=x_{l}$, and $g(x_{l})=x_{l}$ are both identity mapping functions. $g(x_{l})$ expresses the identity mapping of the first-level and second-level shortcuts, and $h(x_{l})$ denotes the identity mapping of the final-level shortcuts. $g(x_{l})$ function is a type B projection shortcut~\cite{he2015resnets}.
\par 
In this paper, we will construct RoR based on ResNets, Pre-ResNets and WRN, in this order. When we use original ResNets as basic residual networks, $f$ is an ReLU function, and RoR adopts a conv-BN-ReLU order in the residual blocks. The architecture of RoR-3 in detail is shown in Fig.~\ref{fig:RoRnetworks}. RoR constructed on Pre-ResNets, and WRN are named after Pre-RoR and RoR-WRN. Their order of residual blocks is BN-ReLU-conv, and all $g(x_{l})$, $h(x_{l})$, and $f$ functions are identity mapping. The architectures of Pre-RoR and RoR-WRN in detail are shown in Fig.~\ref{fig:Pre-RoRnetworks}. The formulations of the three different residual blocks are changed by the following formulas:
\begin{equation}
\begin{array}{cl}
x_{L/3+1}=g(x_{1})+h(x_{L/3})+F(x_{L/3}, W_{L/3})
\end{array}
\label{E:Pre-RoRNets1}
\end{equation}
\begin{equation}
\begin{array}{cl}
x_{2L/3+1}=g(x_{L/3+1})+h(x_{2L/3})+F(x_{2L/3}, W_{2L/3})
\end{array}
\label{E:Pre-RoRNets2}
\end{equation}
\begin{equation}
\begin{array}{cl}
x_{L+1}=g(x_{1})+g(x_{2L/3+1})+h(x_{L})+F(x_{L}, W_{L})
\end{array}
\label{E:Pre-RoRNets3}
\end{equation}
\par
\begin{figure}
\centering
\includegraphics[width=0.8\linewidth]{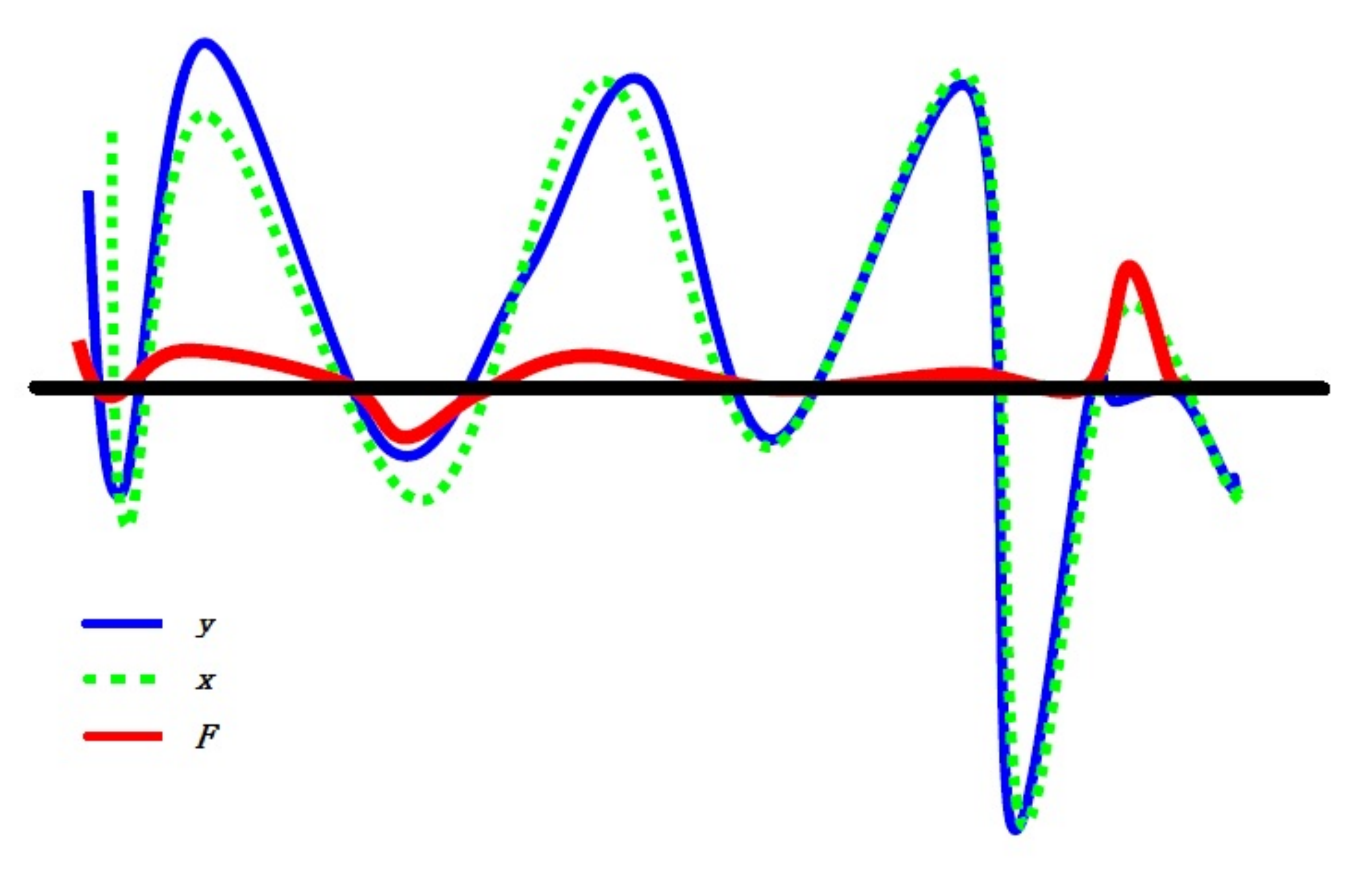}
\caption{Different mapping in residual block of ResNets. Residual networks transfer the learning problem from reaching $y_{l}$ (the blue line) to reaching $F(x_{l}, W_{l})$ (the red line), and we can find the red line is simpler and easier to learn than the blue line. }
\label{fig:curve}
\end{figure}

 
At least two reasons for promoting the optimization ability of RoR by adding extra shortcut connections are presented. First, ResNets transform the learning of $y_{l}$ into the learning of $F(x_{l}, W_{l})$ by residual block structure, since residual mapping function $F$ is easier to learn than $y_{l}$, as shown in Fig.~\ref{fig:curve}. RoR adds extra shortcuts above the original residual blocks, and $y_{l}$ also becomes residual mapping. So RoR transfers the learning problem to learning the residual mapping of residual mapping which is simpler and easier to learn than the original residual mapping. Second, ResNets use shortcuts to propagate information only between neighboring layers in residual blocks. RoR creates several direct paths for propagating information between different original residual blocks by adding extra shortcuts, so layers in upper blocks can propagate information to layers in lower blocks. By information propagation, RoR can alleviate the vanishing gradients problem. The good results of the following experiments show that RoR benefits from the standpoint of optimization through RoR residual mapping and the extra shortcuts to expedite information propagation between layers.
\section{Optimization of RoR}
In order to optimize RoR, we must determine some important parameters and principles, such as shortcut level number, identity mapping type, maximum epoch number and whether to use the drop-path.
\subsection{Shortcut level number of RoR}
\begin{figure}
\centering
\includegraphics[width=1.0\linewidth]{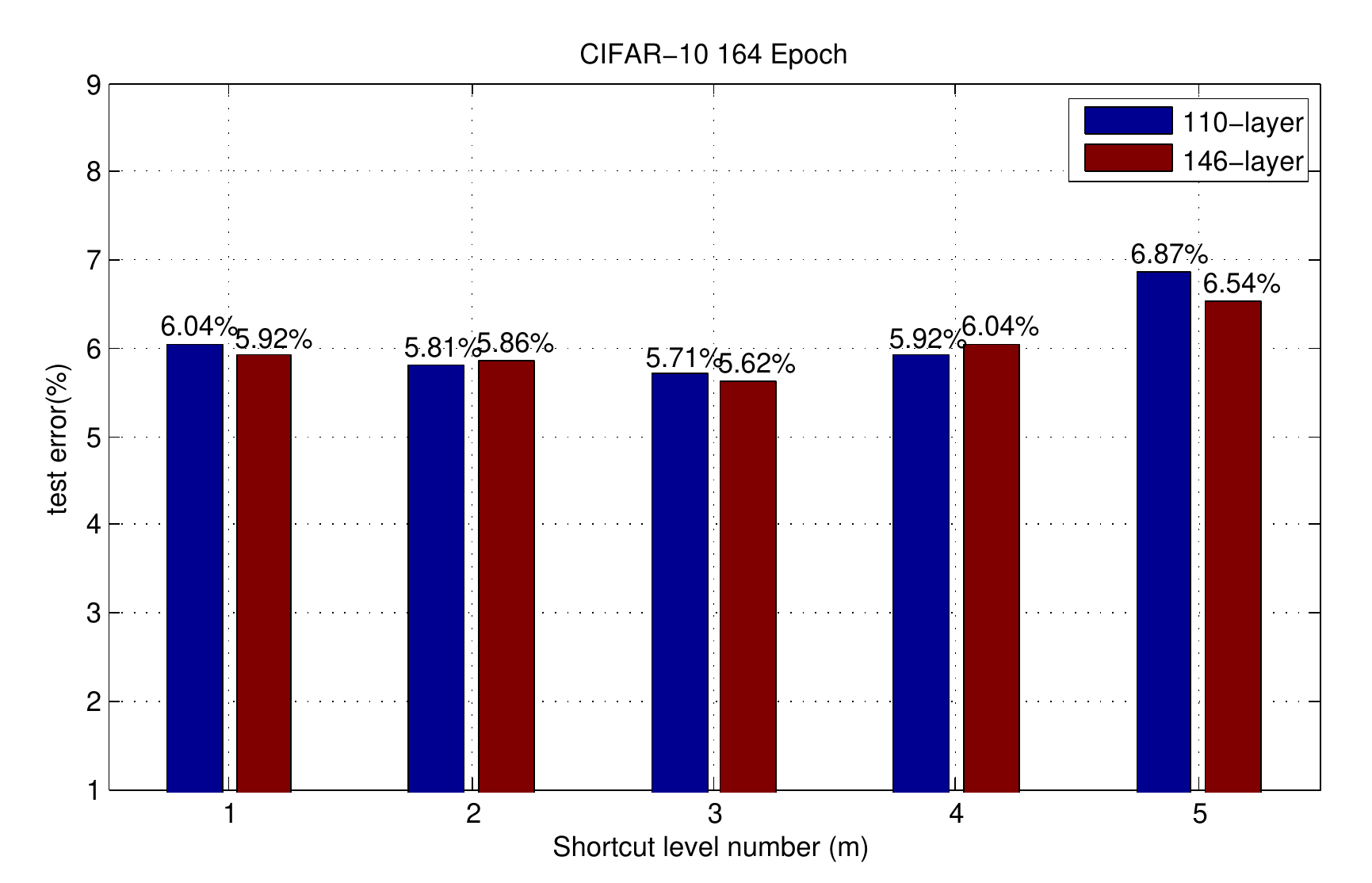}
\caption{Comparison of RoR with different shortcut levels $m$. When $m$=1, it is the original ResNets. When $m$=3, RoR reaches the best performance. }
\label{fig:shortlevel}
\end{figure}

It is important to choose a suitable number of RoR levels for a satisfying performance. The more shortcut levels chosen, the more branches and parameters are added. The overfitting problem will be exacerbated, and the performance may decrease. However, RoR improvements will be less obvious if the number of levels is too small. So we must find a suitable number to keep the balance. We conducted some experiments on CIFAR-10 with different depths and shortcut levels, and the results are described in Fig.~\ref{fig:shortlevel}. RoR ($m$=3) had the best performance of all, and the performance of RoR ($m$=4 or 5) was worse than the original ResNets ($m$=1). So we chose $m$=3, which is shown in Fig.~\ref{fig:RoRnetworks} and denoted as RoR-3.
\subsection{Identity Mapping Types of RoR}
He et al.~\cite{he2015resnets} investigated three types of projection shortcuts: (A) Zero-padding shortcuts are used for increasing dimensions, and all shortcuts are parameter-free; (B) projection shortcuts (done by 1$\times$1 convolutions) are used for increasing dimensions, while other shortcuts are identity connections, and (C) all shortcuts are projections. Type B is slightly better than Type A, and Type C is marginally better than B. But Type C has too many extra parameters, so we used Type A or B as the type of final-level shortcuts. 
\par
\begin{figure}
\centering
\includegraphics[width=0.7\linewidth]{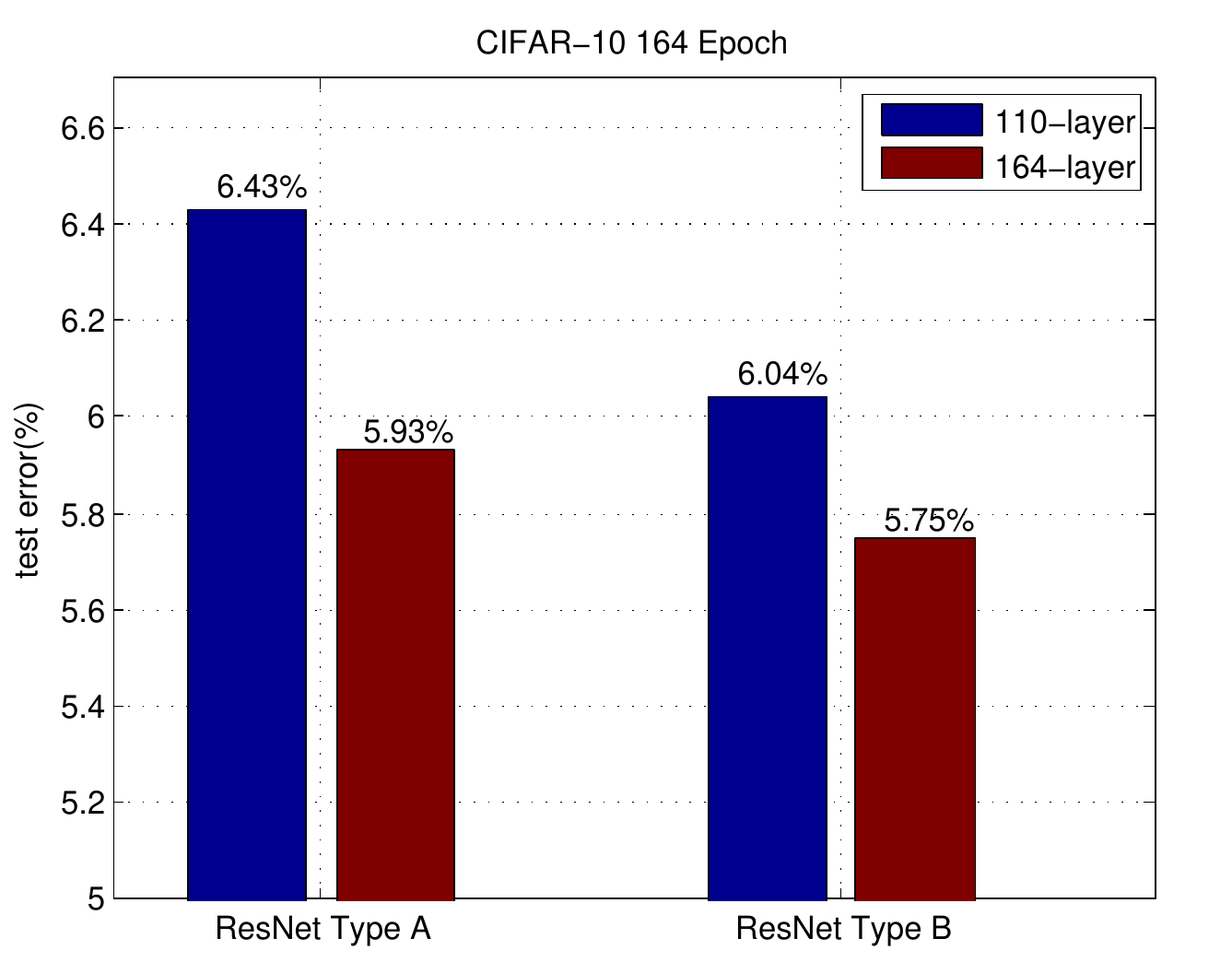}
\caption{Comparison of ResNets with different identity mapping types on CIFAR-10. Using Type B can achieve a better performance than using Type A on CIFAR-10.}
\label{fig:mappingtype10}
\end{figure}

\begin{figure}
\centering
\includegraphics[width=0.7\linewidth]{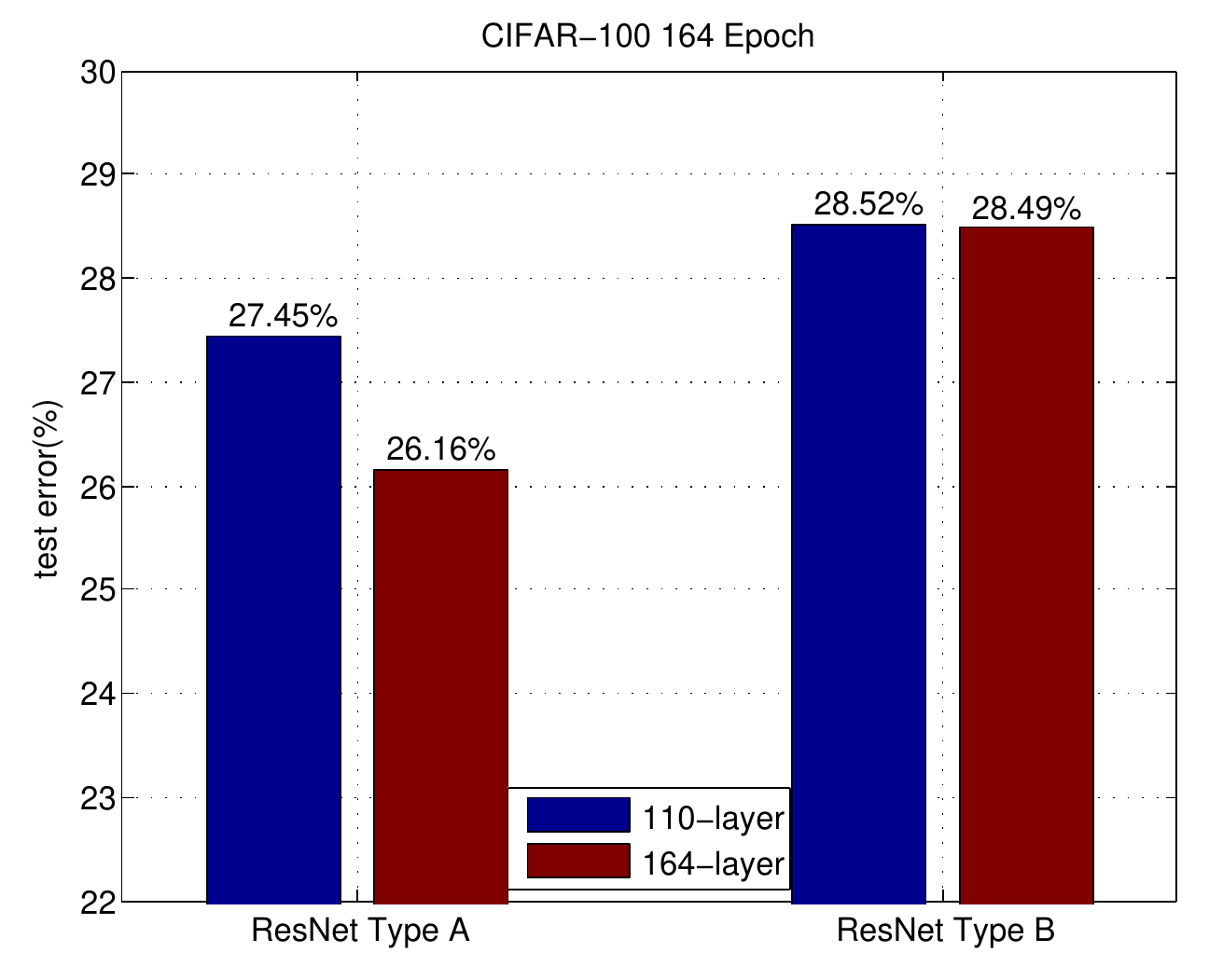}
\caption{Comparison of ResNets with different identity mapping types on CIFAR-100. Using Type A can achieve a better performance than using Type B on CIFAR-100.}
\label{fig:mappingtype100}
\end{figure}

 
Because CIFAR-10 has only 10 classes, the problem of overfitting is not critical, and extra parameters will not obviously escalate overfitting; Thus, we used Type B in the original residual blocks on CIFAR-10. Fig.~\ref{fig:mappingtype10} shows that we can achieve better performance using Type B than Type A on CIFAR-10. However, for CIFAR-100, which has 100 classes with less training examples, overfitting is critical, so we used Type A in the final level. Fig.~\ref{fig:mappingtype100} shows that we can achieve better performance using Type A than Type B on CIFAR-100. The original ResNets were used in all these experiments.
\par 
The shortcuts in Level 1 and Level 2 of RoR are all projection shortcuts. We used Type B in these levels, because the input and output planes of these shortcuts are very different (especially for Level 1), and the zero-padding (Type A) will bring more deviation. Table I shows  the final level where using Type A (on CIFAR-100) or Type B (on CIFAR-10), while other levels using Type B can achieve better performance than pure type A, independent of whether $m$=2 or $m$=3. In Table~\ref{tab:shortcuttype}, Type A and Type B indicate the shortcut type in all but the final level.
\begin{table}[!t]
\renewcommand{\arraystretch}{1.3}
\caption{Test Error (\%) on CIFAR-10/100 with Different Shortcut Type and Level}
\label{tab:shortcuttype}
\centering
\begin{tabular}{|p{1.35cm}|p{0.8cm}|p{0.8cm}|p{0.8cm}|p{0.8cm}|p{0.8cm}|}
\hline
500 Epoch                    &ResNets             &RoR-2 TypeA   &RoR-3 TypeA &RoR-2 TypeB &RoR-3 TypeB \\ \hline\hline
CIFAR-10 110-layer           &5.43                  &6.32          &7.45        &5.22        &5.08        \\\hline
CIFAR-100 110-layer          &26.80               &28.36         &30.12       &27.19       &26.64       \\\hline
\end{tabular}
\end{table}

\par 
In the following Pre-RoR and RoR-WRN experiments, we found the results were comparable whether we used Type B or Type A on CIFAR-10. So in order to keep consistent with shortcut types on CIFAR-100, we all used Type A in the final shortcut level, and Type B in the other shortcut levels. 
\subsection{Maximum Epoch Number of RoR}
\begin{figure}
\centering
\includegraphics[width=0.7\linewidth]{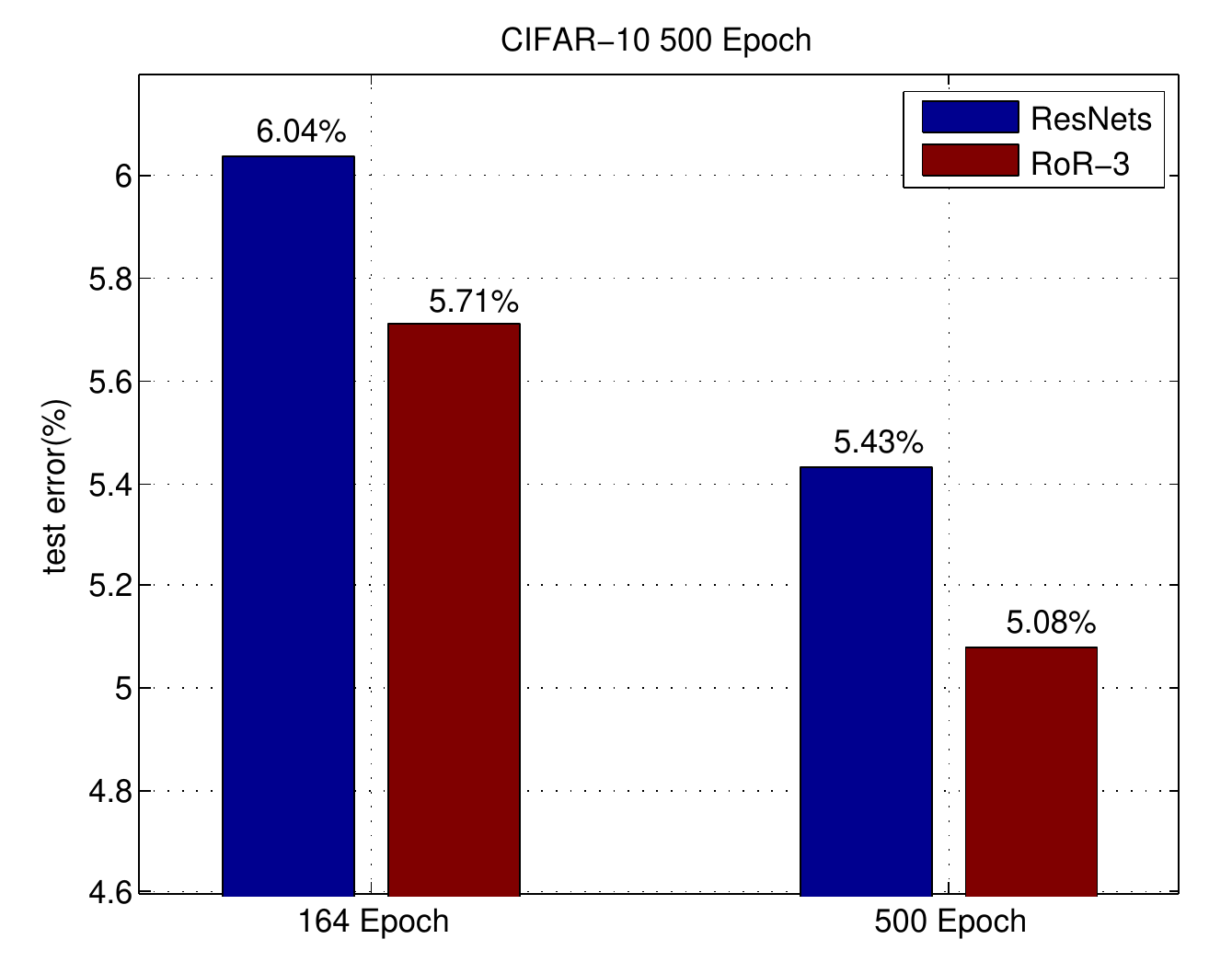}
\caption{Comparison of 110-layer ResNets and RoR-3 with different epoch numbers on CIFAR-10. 500 epochs can achieve better performance than 164 epochs on CIFAR-10.}
\label{fig:cifar10epcohnumber}
\end{figure}

\begin{figure}
\centering
\includegraphics[width=0.7\linewidth]{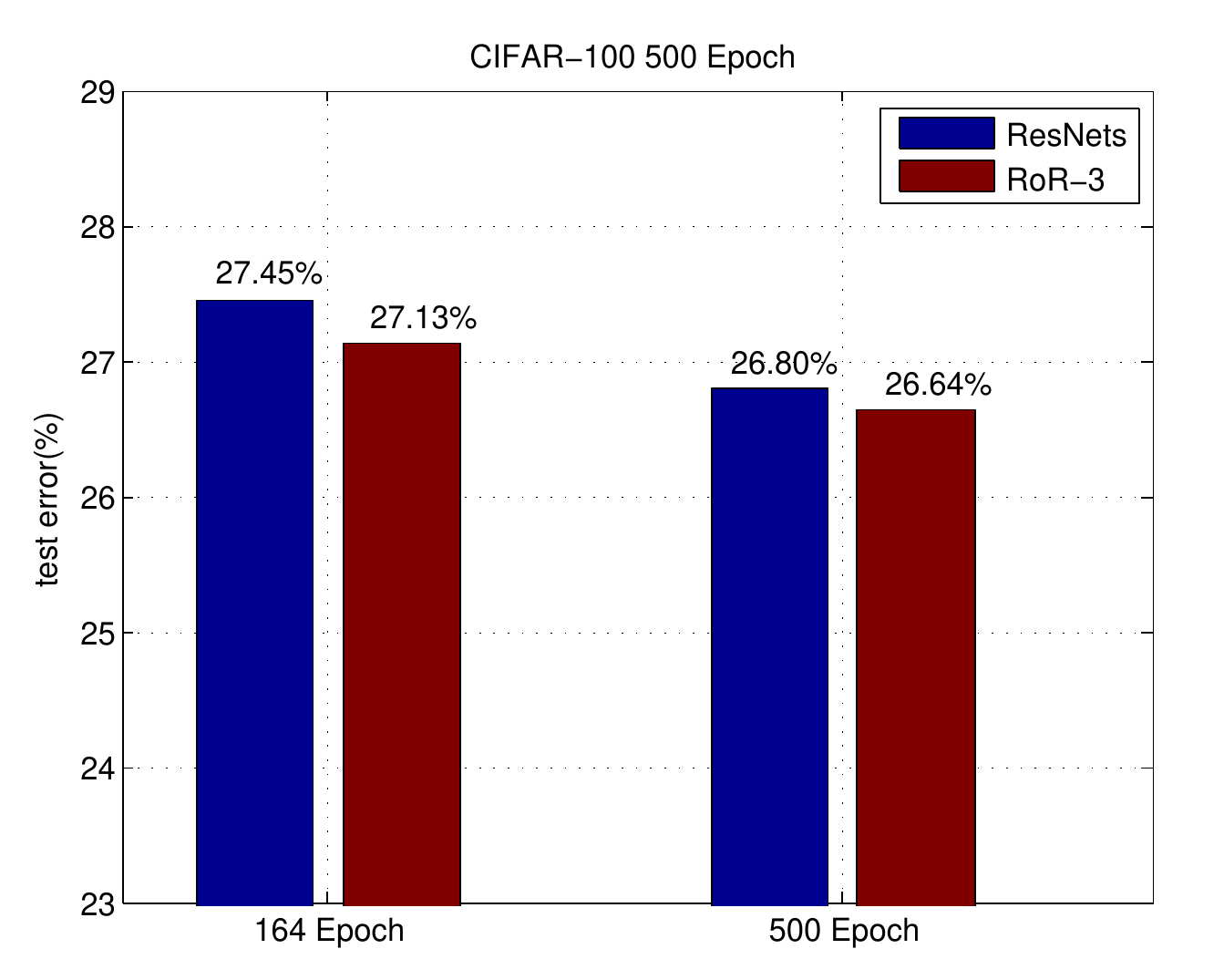}
\caption{Comparison of 110-layer ResNets and RoR-3 with different epoch numbers on CIFAR-100. 500 epochs can achieve better performance than 164 epochs on CIFAR-100.}
\label{fig:cifar100epcohnumber}
\end{figure}

He et al.~\cite{he2015resnets,he2016preresnets} adopted 164 epochs to train CIFAR-10 and CIFAR-100, but we found this epoch number inadequate to optimize ResNets and RoR. Fig.~\ref{fig:cifar10epcohnumber} and Fig.~\ref{fig:cifar100epcohnumber} show that training 500 epochs can get significant promotion. So in this paper, we choose 500 as the maximum epoch number.
\subsection{Drop Path by Stochastic Depth}
Overfitting can be a critical problem for the CIFAR-100 data set. Adding extra shortcuts to the original ResNets can cause the overfitting problems to be even more severe. So our RoR must employ a method to alleviate the overfitting problem. The most frequently used methods are dropout~\cite{hinton2012dropout,sriva2014dropout} and drop-path~\cite{wan2013dropconnect}, which modify interactions between sequential network layers in order to discourage co-adaptaion. Dropout is less effective when used in convolutional layers, and the results of WRN~\cite{zagoruyko2016wrn} also proved that the effect of dropout in residual networks was unapparent. So we did not employ dropout in RoR. Drop-paths prevent co-adaptation of parallel paths by randomly dropping the path. He et al.~\cite{he2016preresnets} proved that the network cannot converge to a good solution by dropping an identity mapping path randomly, because dropping an identity mapping path greatly influences training. However, Huang et al.~\cite{huang2016SD} proposed a stochastic depth drop-path method which only dropped the residual mapping path randomly. Their experiments showed that the method reduced the test errors significantly. So in this paper we use the stochastic depth drop-path method in our RoR except for the ImageNet data set, and it can significantly alleviate overfitting, especially on the CIFAR-100 data set.

\section{Experiments and Analysis}
We empirically demonstrated the effectiveness of RoR on a series of benchmark data sets: CIFAR-10, CIFAR-100, SVHN and ImageNet.
\subsection{Implementation}
For these data sets we compared between the results of RoR and the original ResNets baseline, and other state-of-the-art methods. In the case of CIFAR, we used the same 110-layer and 164-layer ResNets used by~\cite{he2015resnets} to construct RoR architecture. The original ResNets contained three groups of 16 filters, 32 filters and 64 filters of residual blocks, and the feature map sizes are 32, 16 and 8, respectively. The 110-layer RoR contained 18 final residual blocks, three middle-level residual blocks (every middle-level residual block contained six final residual blocks), and one root-level residual block (the root-level residual block contained three middle-level residual blocks). The 164-layer RoR contained 27 final residual blocks, three middle-level residual blocks (every middle-level residual block contained nine final residual blocks), and one root-level residual block. Our implementations are based on Torch 7 with one Nvidia Geforce Titan X. We adopted batch normalization (BN)~\cite{ioffe2015bn} after each convolution in residual mapping paths and before activation (ReLU)~\cite{nair2010relu}, as shown in Fig.~\ref{fig:RoRnetworks}. In Pre-RoR and RoR-WRN experiments, we adopted BN-ReLU-conv order, as shown in Fig.~\ref{fig:Pre-RoRnetworks}. We initialized the weights as in~\cite{he2015prelu}. For CIFAR data sets, we used SGD with a mini-batch size of 128 for 500 epochs. The learning rate starts from 0.1, and is divided by a factor of 10 after epoch 250 and 375 as in~\cite{huang2016SD}. For the SVHN data set, we used SGD with a mini-batch size of 128 for 50 epochs. The learning rate starts from 0.1, and it is divided by a factor of 10 after epoch 30 and 35 as in~\cite{huang2016SD}. We used a weight decay of 1e-4, a momentum of 0.9, and a Nesterov momentum with 0 dampening on all data sets~\cite{gross2016facebookres}. For the stochastic depth drop-path method, we set $p_{l}$ with the linear decay rule of $p_{0}=1$ and $p_{L}$=0.5~\cite{huang2016SD}. 
\subsection{CIFAR-10 Classification by RoR}
CIFAR-10 is a data set of 60,000 32$\times$32 color images, with 10 classes of natural scene objects. The training set and test set contain 50,000 and 10,000 images. Two standard data augmentation techniques~\cite{he2015resnets,he2016preresnets,zagoruyko2016wrn,huang2016SD} were adopted in our experiments: random sampling and horizontal flipping. We preprocessed the data by subtracting the mean and dividing the standard deviation.
\begin{table}[!t]
\renewcommand{\arraystretch}{1.3}
\caption{Test Error (\%) on CIFAR-10 ResNets and RoR}
\label{tab:cifar10RoR}
\centering
\begin{tabular}{|l|c|c|c|c|}
\hline
CIFAR-10 500 Epoch           &ResNets             &ResNets+SD  &RoR-3    &RoR-3+SD  \\ \hline\hline
110-layer                    &5.43                &5.63        &5.08     &5.04    \\\hline
164-layer                    &5.07                &5.06        &4.86     &4.90    \\\hline
\end{tabular}
\end{table}
\par
\begin{figure}
\centering
\includegraphics[width=1.0\linewidth]{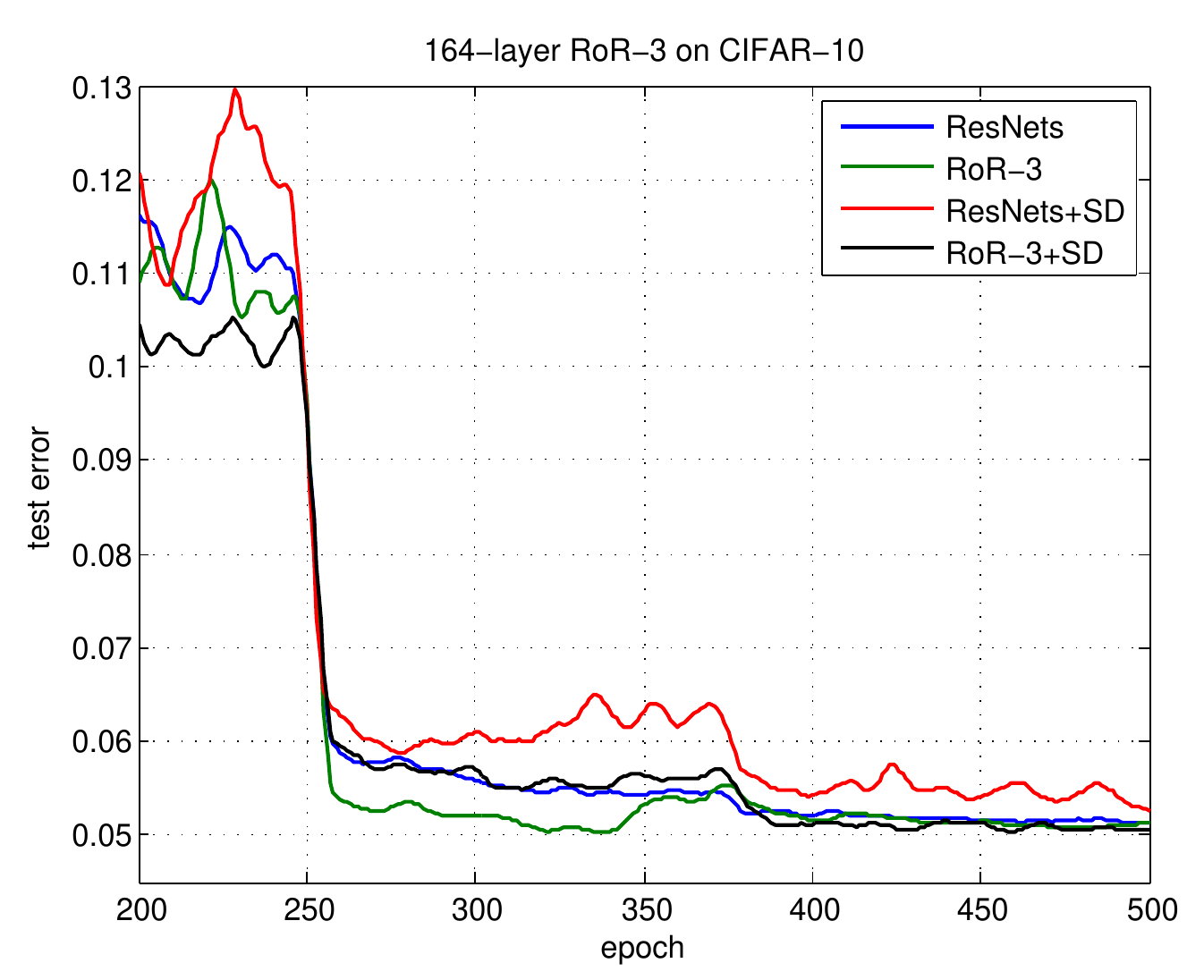}
\caption{Smoothed test errors on CIFAR-10 by ResNets, RoR-3, ResNets+SD and RoR-3+SD during training, corresponding to results in Table~\ref{tab:cifar10RoR}. Either RoR-3 without SD (the green curve) or RoR-3+SD (the black curve) is shown yielding a lower test error than ResNets. }
\label{fig:RoR-310}
\end{figure}

 
In Table~\ref{tab:cifar10RoR} and Fig.~\ref{fig:RoR-310}, the 110-layer ResNets without SD resulted in a competitive 5.43\% error on the test set. The 110-layer RoR-3 without SD had a 5.08\% error on the test set and outperformed the 110-layer ResNets without SD by 6.4\% (Because all state-of-the-art methods have achieved similarly small error rates, we used relative percentage to measure the improvements in this paper) on CIFAR-10 with a similar number of parameters. The 164-layer RoR-3 without SD resulted in a 4.86\% error on the test set, and it outperformed the 164-layer ResNets without SD by 4.1\%. As can be observed, the 164-layer RoR-3 without SD can also outperform the 4.92\% error of the 1001-layer Pre-ResNets with the same mini-batch size~\cite{he2016preresnets}. We then added SD on ResNets and RoR-3, but those performances were similar to the models without SD. We concluded that overfitting on CIFAR-10 is not critical, and SD is not effective. However, adding SD can reduce training time~\cite{huang2016SD} and does not affect the performance, so we added SD in the following experiments on CIFAR-10.
\subsection{CIFAR-100 Classification by RoR}
Similar to CIFAR-10, CIFAR-100 is a data set of 60,000 32$\times$32 color images, but with 100 classes of natural scene objects. The training set and test set contain 50,000 and 10,000 images, respectively. The augmentation and preprocessing techniques adopted in our experiments are the same as on CIFAR-10.
\begin{table}[!t]
\renewcommand{\arraystretch}{1.3}
\caption{Test Error (\%) on CIFAR-100 ResNets and RoR}
\label{tab:cifar100RoR}
\centering
\begin{tabular}{|p{1.35cm}|p{0.7cm}|p{0.8cm}|p{0.7cm}|p{0.8cm}|p{0.7cm}|p{0.8cm}|}
\hline
CIFAR-100 500 Epoch          &ResNets             &ResNets +SD  &RoR-2     &RoR-2 +SD  &RoR-3     &RoR-3 +SD  \\ \hline\hline
110-layer                    &26.80               &23.83       &27.19     &23.60     &26.64     &23.48   \\\hline
164-layer                    &25.85               &23.29       &-         &-         &27.45     &22.47 \\\hline
\end{tabular}
\end{table}
\par
\begin{figure}
\centering
\includegraphics[width=1.0\linewidth]{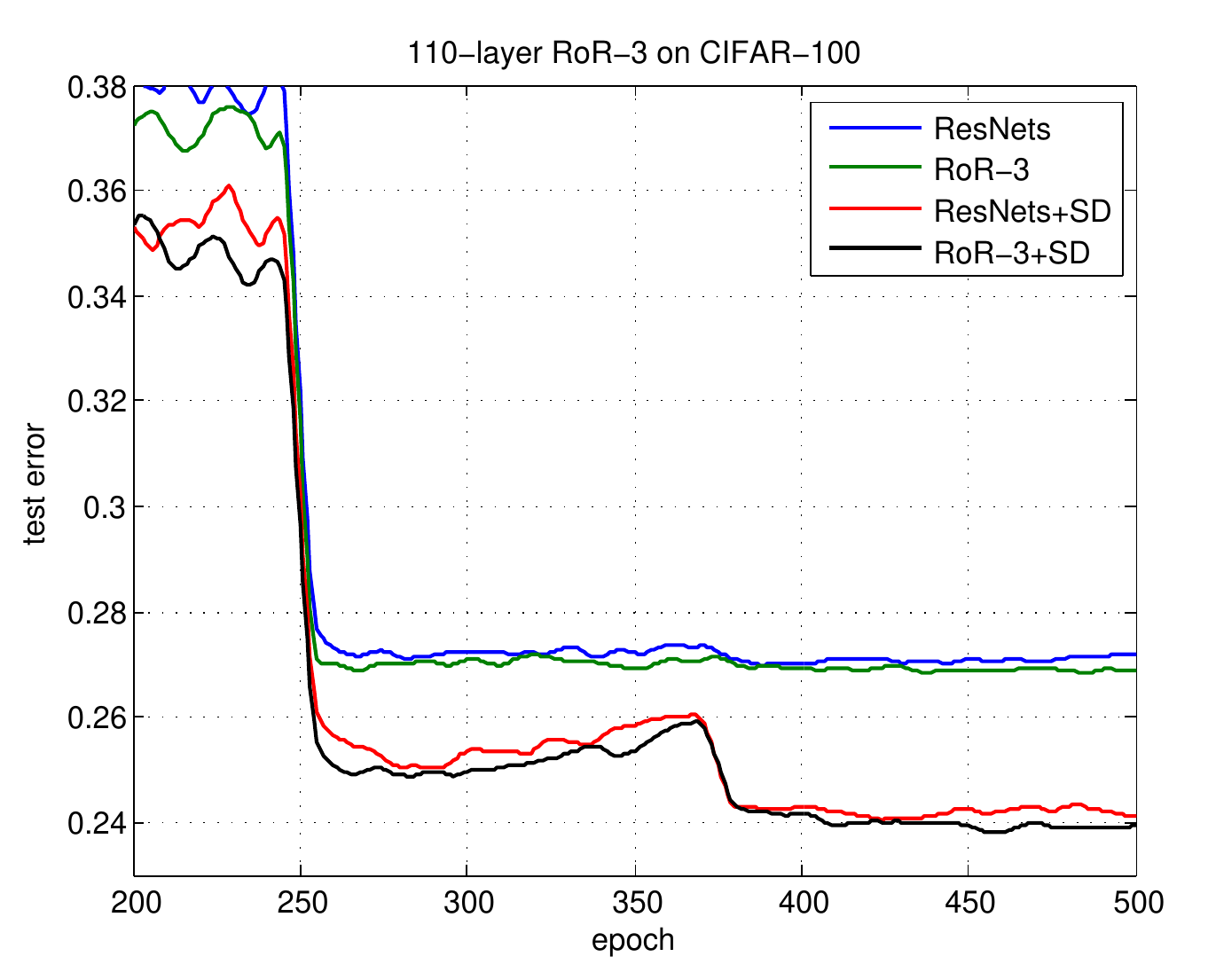}
\caption{Smoothed test error on CIFAR-100 by ResNets, RoR-3, ResNets+SD and RoR-3+SD during training, corresponding to results in Table~\ref{tab:cifar100RoR}. RoR-3+SD (the black curve) yields lower test errors than other curves.}
\label{fig:RoR-3100}
\end{figure}

 
In Table~\ref{tab:cifar100RoR} and Fig.~\ref{fig:RoR-3100}, the 110-layer and 164-layer ResNets without SD resulted in a competitive 26.80\% and 25.85\% error on the test set, but the results of the 110-layer RoR-3 and 164-layer RoR-3 without SD were not ideal. We argue that this is because adding extra branches and convolutional layers may escalate overfitting. It is gratifying that the 110-layer RoR-3+SD and 164-layer RoR-3+SD resulted in a 23.48\% and 22.47\% error on the test set, and they outperformed the 110-layer ResNets, 110-layer ResNets+SD, 164-layer ResNets and 164-layer ResNes+SD by 12.4\%, 1.5\%, 13.1\% and 3.5\%, respectively on CIFAR-100. This indicates that the SD drop-path can alleviate overfitting, so we will add SD in the next experiments on CIFAR-100. In addition, we observe that the optimization ability of RoR-2 is better than original ResNets, but worse than RoR-3, and that is why we chose $m$=3.
\subsection{Residual Block Size Analysis}
In the above experiments, we used the residual block with two 3$\times$3 convolution layers B(3,3)~\cite{he2015resnets}. In order to analyze the effects of different residual block sizes, we increased the convolution layer number of every residual block to three with the same total number of layers, and the new residual block was denoted by B(3,3,3). The results are shown in Table~\ref{tab:DFBlock}. When the epoch number was 164, we achieved better performance by B(3,3,3). But if the epoch number is 500 (RoR fully trained), we found the results by B(3,3) to be better than those obtained by B(3,3,3). WRN~\cite{zagoruyko2016wrn} tried more kinds of residual blocks, and B(3,3) remained the best residual block type and size. So we chose B(3,3) as the basic residual block size in RoR. Again, the importance of 500 epochs and $m$=3 was validated.
\begin{table}[!t]
\renewcommand{\arraystretch}{1.3}
\caption{Test Error (\%) on CIFAR-10 with Different Block Size}
\label{tab:DFBlock}
\centering
\begin{tabular}{|p{1.5cm}|p{1.3cm}|p{1.3cm}|p{1.3cm}|p{1.3cm}|}
\hline
CIFAR-10                     &RoR-3 B(3,3)        &RoR-3 B(3,3,3)  &RoR-4 B(3,3)    &RoR-4 B(3,3,3)  \\ \hline\hline
164-layer 164 Epoch          &6.34                &5.77            &5.94            &5.21    \\\hline
164-layer 500 Epoch          &4.86                &5.12            &5.09            &5.20    \\\hline
\end{tabular}
\end{table}
\par 
\subsection{Versatility of RoR for other residual networks}
Recently several variants of residual networks have become available, which can improve the performance of original ResNets~\cite{he2015resnets}. For example, Pre-ResNets~\cite{he2016preresnets} can reduce vanishing gradients by BN-ReLU-conv order, and WRN~\cite{zagoruyko2016wrn} can achieve a dramatic performance increase by adding more feature planes based on Pre-ResNets. In this paper, we constructed the RoR architecture based on these two residual networks. 
\par 
First, we changed the residual blocks of the original RoR with a BN-ReLU-conv order, which can only be done by adding two-level shortcuts on the Pre-ResNets. Fig.~\ref{fig:Pre-RoRnetworks} shows the architecture of Pre-RoR ($k$=1) in detail. We did the same experiment by Pre-RoR-3 on CIFAR-10 and CIFAR-100, and the results are shown in Table~\ref{tab:Pre-RoR} where Pre-RoR is compared with Pre-ResNets. As can be observed, the 164-layer Pre-RoR-3+SD had a surprising 4.51\% error on CIFAR-10 and a 21.94\% error on CIFAR-100. Particularly, the  164-layer Pre-RoR-3+3D outperformed the 164-layer Pre-ResNets and 164-layer Pre-ResNets+SD by 14.1\% and 2.4\% on CIFAR-100. 
\begin{table}[!t]
\renewcommand{\arraystretch}{1.3}
\caption{Test Error (\%) on CIFAR-10/100 by Pre-ResNets and Pre-RoR}
\label{tab:Pre-RoR}
\centering
\begin{tabular}{|p{1.5cm}|p{1.3cm}|p{1.3cm}|p{1.3cm}|p{1.3cm}|}
\hline
500 Epoch                     &Pre-ResNets         &Pre-RoR-3       &Pre-ResNest+SD  &Pre-RoR-3+SD  \\ \hline\hline
164-layer CIFAR-10            &5.04                &5.02            &4.67            &4.51    \\\hline
164-layer CIFAR-100           &25.54               &25.33           &22.49           &21.94    \\\hline
\end{tabular}
\end{table}
\par
\begin{figure}
\centering
\includegraphics[width=1.0\linewidth]{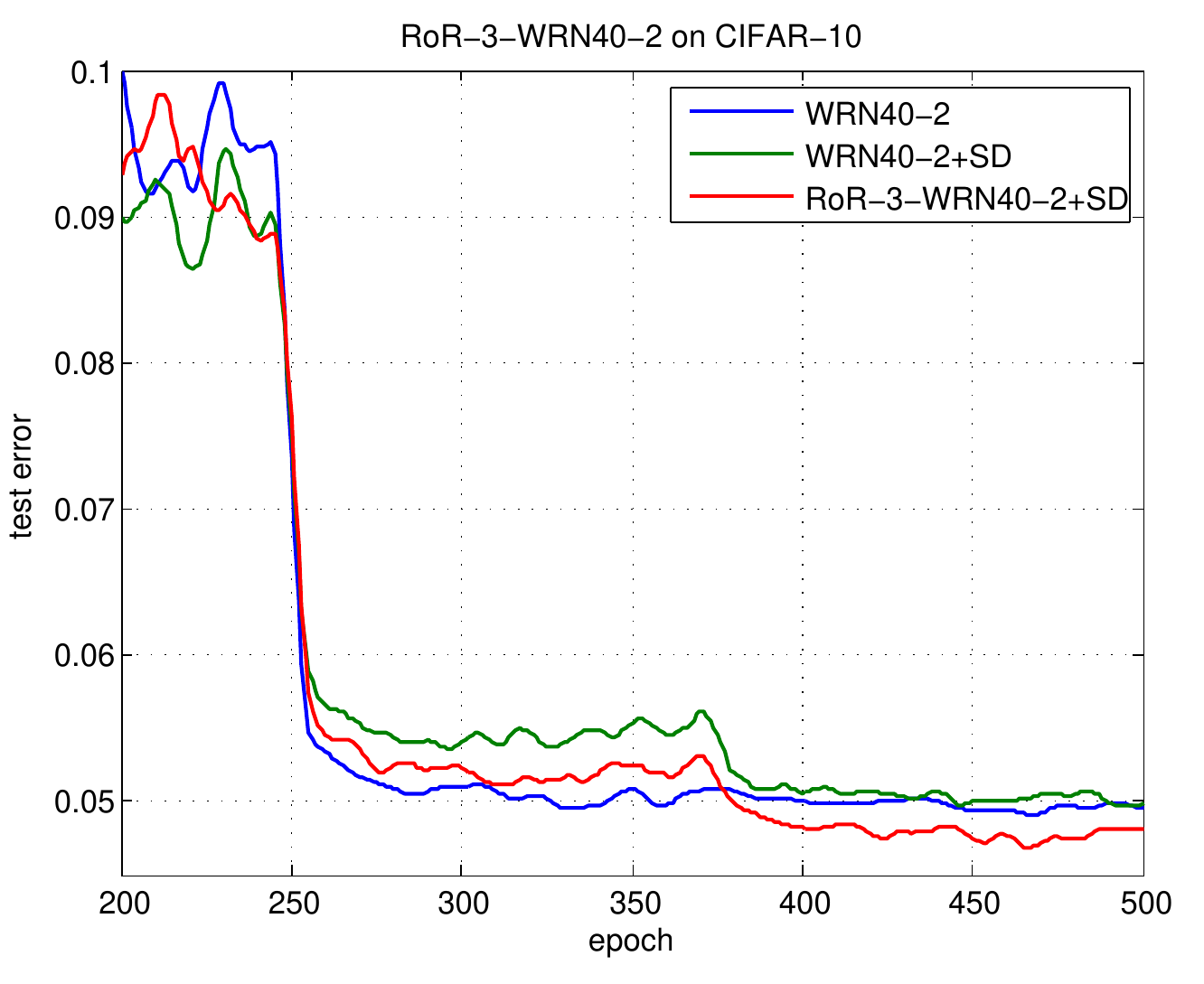}
\caption{Smoothed test error on CIFAR-10 by WRN40-2, WRN40-2+SD and RoR-3-WRN40-2+SD during training, corresponding to results in Table~\ref{tab:RoR-WRN}. RoR-3-WRN40-2+SD (the red curve) yields lower test errors than the other curves.}
\label{fig:WRN10}
\end{figure}
\begin{figure}
\centering
\includegraphics[width=1.0\linewidth]{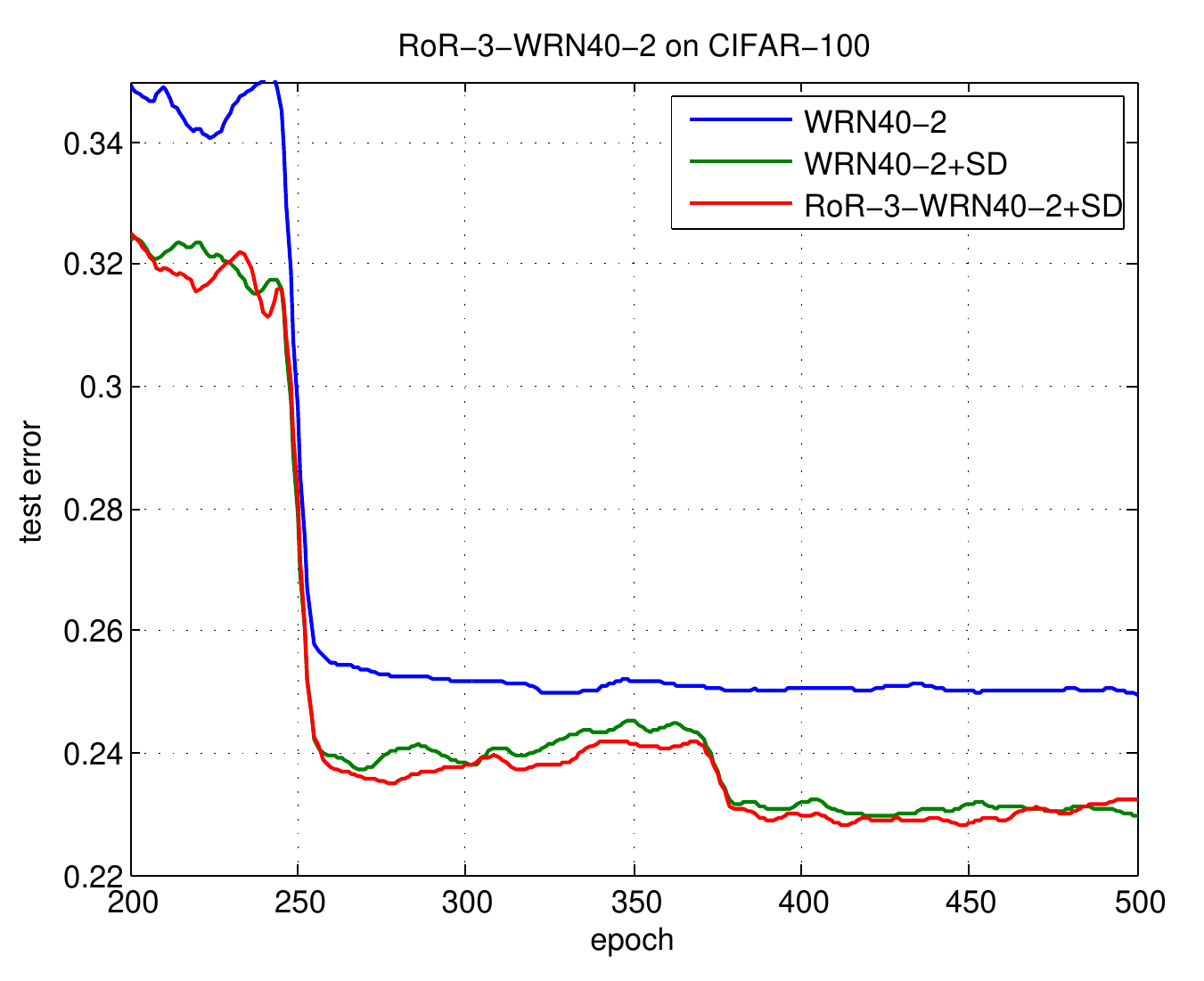}
\caption{Smoothed test error on CIFAR-100 by WRN40-2, WRN40-2+SD and RoR-3-WRN40-2+SD during training, corresponding to results in Table~\ref{tab:RoR-WRN}. RoR-3-WRN40-2+SD (the red curve) yields lower test errors than the other curves.}
\label{fig:WRN100}
\end{figure}

Second, we used (16$\times$k, 32$\times$k, 64$\times$k) filters instead of (16, 32, 64) filters of the original Pre-RoR because WRN is constructed based on Pre-ResNets. Fig.~\ref{fig:Pre-RoRnetworks} shows the architecture of RoR-WRN ($k$=2, 4, in detail. We did the same experiment by RoR-3-WRN on CIFAR-10 and CIFAR-100 and showed the results in Table~\ref{tab:RoR-WRN}. Fig.~\ref{fig:WRN10} and Fig.~\ref{fig:WRN100} showed the test errors on CIFAR-10 and CIFAR-100 at different training epochs. As can be observed, the performance of RoR-3-WRN is worse than WRN. In our opinion, WRN has more feature planes, so it is easier to get overfitting when we add extra branches and parameters. SD can alleviate overfitting, so the performance of RoR-3-WRN+SD is always better than others. RoR-3-WRN40-2+SD achieved 4.59\% error on CIFAR-10 and 22.48\% error on CIFAR-100.
\begin{table}[!t]
\renewcommand{\arraystretch}{1.3}
\caption{Test Error (\%) on CIFAR-10/100 by WRN and RoR-WRN}
\label{tab:RoR-WRN}
\centering
\begin{tabular}{|p{1.5cm}|p{1.3cm}|p{1.3cm}|p{1.3cm}|p{1.3cm}|}
\hline
500 Epoch           &WRN40-2             &RoR-3-WRN40-2       &WRN40-2+SD  &RoR-3-WRN40-2+SD  \\ \hline\hline
CIFAR-10            &4.81                &5.01                &4.80        &4.59    \\\hline
CIFAR-100           &24.70               &25.19               &22.87       &22.48    \\\hline
\end{tabular}
\end{table}
\par 
Through analysis and experiments, we can prove that our RoR architecture can also promote optimization abilities of other residual networks, such as Pre-ResNets and WRN. Because RoR has good versatility for other residual networks, we have reasons to believe that our RoR is a good application prospect for the residual-networks family.
\subsection{Depth and Width Analysis}
According to preceding experiments in this section, we determined that performance can be improved by increasing depth or width. In this section we analyze how to choose depth and width of RoR.
\par 
The basic RoR is based on the original ResNets, but very deep ResNets encounter serious vanishing gradients problems. So even though the performance of RoR is better than ResNets, RoR still cannot resolve the vanishing gradients problem. We repeated the RoR experiments by increasing the number of convolutional layers, as shown in Table~\ref{tab:RoRDFDepth}. As can be observed, when the number of layers increased from 164 to 182, and then to 218, the performance gradually decreased. These experiments demonstrated that the vanishing problem exists in very deep RoR.
\begin{table}[!t]
\renewcommand{\arraystretch}{1.3}
\caption{Test Error (\%) on CIFAR-10/100 by RoR with Different Depths}
\label{tab:RoRDFDepth}
\centering
\begin{tabular}{|l|c|c|}
\hline
Depth           &CIFAR-10 RoR-3 without SD             &CIFAR-100 RoR-3+SD  \\ \hline\hline
110-layer       &5.08                                  &23.48               \\\hline
164-layer       &4.86                                  &22.47               \\\hline
182-layer       &4.98                                  &22.76               \\\hline
218-layer       &5.12                                  &22.99               \\\hline
\end{tabular}
\end{table}
\par 
Pre-ResNets reduced vanishing problem, so Pre-RoR should inherit this property too. We repeated the Pre-RoR experiments by increasing the number of convolutional layers, as shown in Table~\ref{tab:Pre-RoRDFDepth}. We observed that the accuracy increased as the number of layers increased. The 1202-layer Pre-RoR-3+SD with a mini-batch size of 32 achieved the 4.49\% error on CIFAR-10 and 20.64\% error on CIFAR-100. These results mean that the vanishing gradients can be reduced, even on very deep models. So we can use Pre-RoR to push the depth limit.
\begin{table}[!t]
\renewcommand{\arraystretch}{1.3}
\caption{Test Error (\%) on CIFAR-10/100 by Pre-RoR with Different Depths}
\label{tab:Pre-RoRDFDepth}
\centering
\begin{tabular}{|p{2.4cm}|p{2.3cm}|p{2.3cm}|}
\hline
Depth           &CIFAR-10 Pre-RoR-3+SD             &CIFAR-100 Pre-RoR-3+SD  \\ \hline\hline
110-layer       &4.63                                  &23.05               \\\hline
164-layer       &4.51                                  &21.94               \\\hline
218-layer       &4.51                                  &21.43               \\\hline
1202-layer with 32 mini-batch size       &4.49                                  &20.64               \\\hline
\end{tabular}
\end{table}
\par 
WRN is not very deep, so the vanishing problem is not obvious. However, overfitting can become severe because of adding more feature planes and parameters. We followed the same protocol with RoR-WRN with different depths and widths on CIFAR-10 and CIFAR-100, as shown in Table~\ref{tab:RoRWRNDFDepth}. We found both deepening and widening the network can improve the performance. But when we widened the RoR-WRN, weight parameters increased exponentially. So we had to complement RoR-WRN by SD to reduce overfitting. As can be observed, RoR-3-WRN-58-4+SD achieved an extraordinary 3.77\% error on CIFAR-10 and a 19.73\% error on CIFAR-100. We found that the RoR-WRN with similar order of magnitude parameters was more effective than Pre-RoR, because the vanishing problem already existed in very deep Pre-RoR.
\begin{table}[!t]
\renewcommand{\arraystretch}{1.3}
\caption{Test Error (\%) on CIFAR-10/100 by RoR-WRN with Different Depths and Widths}
\label{tab:RoRWRNDFDepth}
\centering
\begin{tabular}{|p{2.4cm}|p{2.3cm}|p{2.3cm}|}
\hline
Depth and Width           &CIFAR-10 RoR-3-WRN+SD             &CIFAR-100 RoR-3-WRN+SD  \\ \hline\hline
RoR-3-WRN40-2       &4.59                                  &22.48               \\\hline
RoR-3-WRN40-4       &4.09                                  &22.11               \\\hline
RoR-3-WRN58-2       &4.23                                  &21.50               \\\hline
RoR-3-WRN58-4       &3.77                                  &19.73               \\\hline
\end{tabular}
\end{table}
\par 
Through experiments and analysis, we determined that the depth and width of RoR are equally important to model learning capability. We must carefully choose suitable depth and width on each given task to achieve satisfying results. In this paper we proposed a two-step strategy to choose depths and widths. The first step is to increase the depth of RoR gradually until the performance was saturated. Then increased the width of RoR gradually until the best results were achieved. 
\subsection{Training time comparison on CIFAR-10/100}
We compared the training time of the ResNets-110, RoR-3-110 and RoR-3-110+SD on CIFAR-10/100, as shown in Table~\ref{tab:time}. The training times of ResNets-110 and RoR-3-110 were similar, so RoR did not add more extra training time than original residual networks. In addition, from Table~\ref{tab:time} we got the same conclusion that stochastic depth consistently gave about 25\% speedup in ~\cite{huang2016SD}.
\begin{table}[!t]
\renewcommand{\arraystretch}{1.3}
\caption{Training Time Comparison on CIFAR-10/100}
\label{tab:time}
\centering
\begin{tabular}{|l|c|c|c|}
\hline
Method              &CIFAR-10            &CIFAR-100       \\ \hline\hline
ResNet-110          &9h40m                &9h43m                      \\\hline
RoR-3-110           &9h47m                &9h51m                        \\\hline
RoR-3-110+SD                   &7h43m                &7h45m                \\\hline
\end{tabular}
\end{table}
\subsection{SVHN Classification by RoR}
\begin{figure}
\centering
\includegraphics[width=1.0\linewidth]{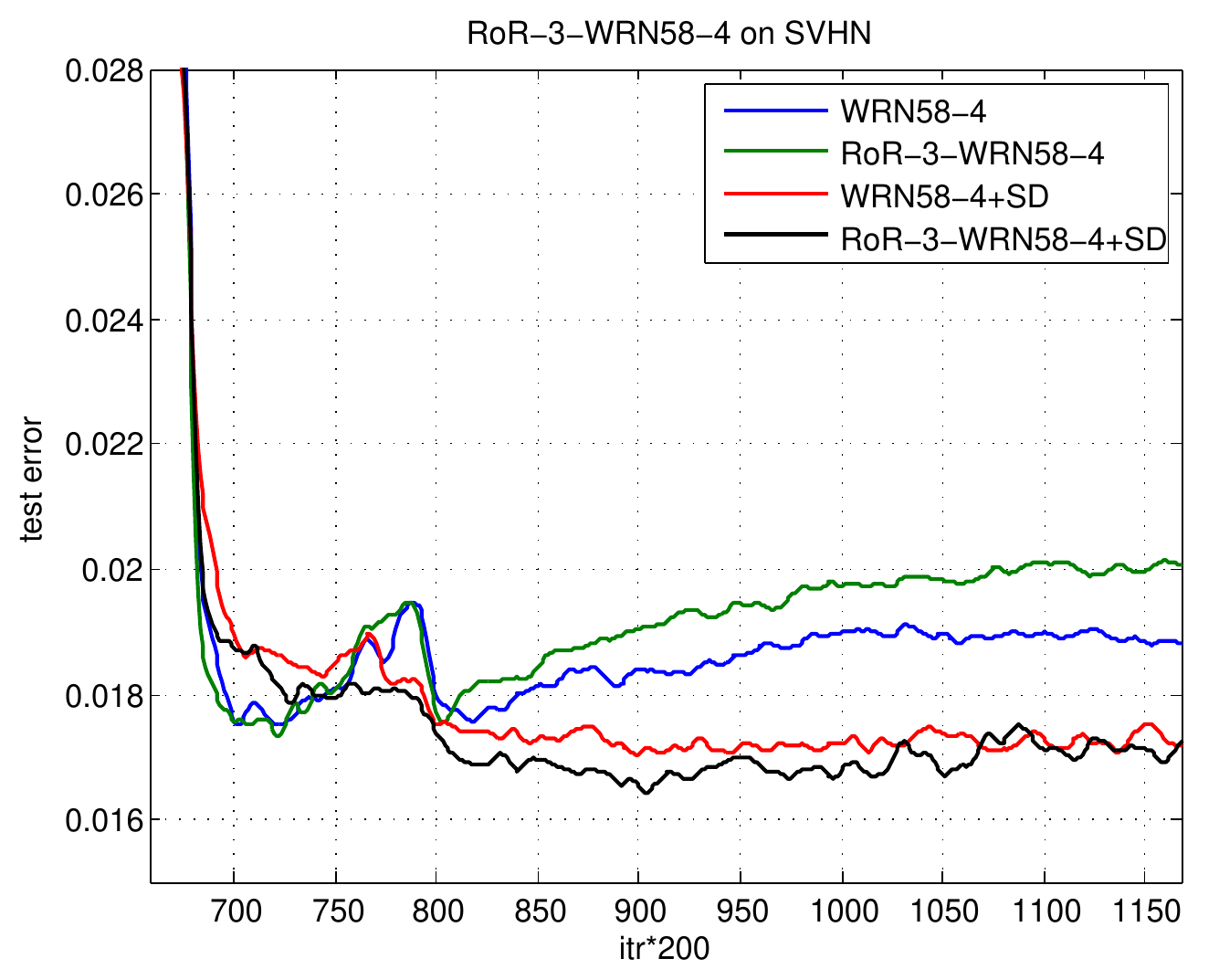}
\caption{Smoothed test error on SVHN by WRN58-4，RoR-3-WRN58-4, WRN58-4+SD and RoR-3-WRN58-4+SD during training. RoR-3-WRN58-4+SD (the black curve) yields lower test errors than the curves.}
\label{fig:SVHN}
\end{figure}

The Street View House Number (SVHN) data set used in this research contains 32$\times$32 colored images of cropped out house numbers from Google Street View. The task is to classify the digit at center (and ignore any additional digit that might appear on the side) of the images. There are 73,257 digits in the training set, 26,032 in the test set and 531,131 easier samples for additional training. Following the common practice, we used all the training samples but did not perform data augmentation. We preprocessed the data by subtracting the mean and dividing the standard deviation. Batch size was set to 128, and test error was calculated every 200 iterations. We used our best architecture RoR-3-WRN58-4+SD to train SVHN and achieved the excellent result of 1.59\% test error, as shown in Table~\ref{tab:SVHN}. This result outperformed WRN58-4 and WRN58-4+SD by 5.9\% and 4.2\% on SVHN, respectively. Fig.~\ref{fig:SVHN} shows the test error at different training epochs. We could see that the results of WRN58-4 and RoR-3-WRN58-4 were also good, but they started to overfit after 700$\times$200 iterations.
\begin{table}[!t]
\renewcommand{\arraystretch}{1.3}
\caption{Test Error (\%) on SVHN by WRN58-4, WRN58-4+SD, WRN58-4+SD and RoR-3-WRN58-4+SD}
\label{tab:SVHN}
\centering
\begin{tabular}{|p{1.1cm}|p{1.3cm}|p{1.3cm}|p{1.3cm}|p{1.7cm}|}
\hline
50 Epoch           &WRN58-4             &RoR-3-WRN58-4      &WRN58-4+SD  &RoR-3-WRN58-4+SD  \\ \hline\hline
SVHN               &1.69                &1.66                &1.66        &1.59    \\\hline
\end{tabular}
\end{table}

\subsection{Comparisons with state-of-the-art results on CIFAR-10/100 and SVHN}
\begin{table}[!t]
\renewcommand{\arraystretch}{1.3}
\caption{Test Error (\%) on CIFAR-10, CIFAR-100 and SVHN by Different Methods}
\label{tab:DFMethods}
\centering
\begin{tabular}{|p{3.5cm}|p{1.2cm}|p{1.2cm}|p{1.2cm}|}
\hline
Method(\#Parameters)           &CIFAR-10             &CIFAR-100      &SVHN \\ \hline\hline
NIN~\cite{lin2013NiN}                          &8.81                 &35.68          &2.35               \\\hline
FitNet~\cite{romero2014fitnets}                       &8.39                 &35.04          &2.42               \\\hline
DSN~\cite{lee2015dsn}                           &7.97                 &34.57          &1.92               \\\hline
All-CNN~\cite{springgenberg2014allcnn}                       &7.25                 &33.71          &-               \\\hline
Highway~\cite{sriva2015highway}                       &7.72                 &32.39          &-               \\\hline
ELU~\cite{clevert2015elu}                           &6.55                 &24.28          &-               \\\hline
FractalNet (30M)~\cite{larsson2016fractalnet}                    &4.59                 &22.85          &1.87               \\\hline\hline
ResNets-164 (2.5M)~\cite{he2015resnets} (reported by~\cite{he2016preresnets}) &5.93             &25.16          &-               \\\hline
FitResNet, LSUV~\cite{mishkin2015initial}               &5.84                 &27.66          &-               \\\hline
Pre-ResNets-164 (2.5M)~\cite{he2016preresnets}        &5.46                 &24.33          &-               \\\hline
Pre-ResNets-1001 (10.2M)~\cite{he2016preresnets}      &4.62                 &22.71          &-               \\\hline
ELU-ResNets-110 (1.7M)~\cite{shah2016elu}        &5.62                 &26.55          &-               \\\hline
PELU-ResNets-110 (1.7M)~\cite{trottier2016pelu}       &5.37                 &25.04          &-               \\\hline
ResNets-110+SD (1.7M)~\cite{huang2016SD}         &5.23                 &24.58          &1.75 (152-layer)               \\\hline
ResNet in ResNet (10.3M)~\cite{targ2016rir}      &5.01                 &22.90          &-               \\\hline
SwapOut (7.4M)~\cite{singh2016swapout}                &4.76                 &22.72          &-               \\\hline
WResNet-d (19.3M)~\cite{shen2016wresnets}             &4.70                 &-              &-               \\\hline
WRN28-10 (36.5M)~\cite{zagoruyko2016wrn}              &4.17                 &20.50          &1.64               \\\hline
CRMN-28 (more than 40M)~\cite{moniz2016crmn}       &4.65                 &20.35          &1.68              \\\hline\hline
RoR-3-164 (2.5M)              &4.86                 &22.47(+SD)          &-               \\\hline
Pre-RoR-3-164+SD (2.5M)          &4.51                 &21.94          &-               \\\hline
RoR-3-WRN40-2+SD (2.2M)          &4.59                 &22.48          &-               \\\hline
Pre-RoR-3-1202+SD (19.4M)        &4.49                 &20.64          &-               \\\hline
RoR-3-WRN40-4+SD (8.9M)          &4.09                 &20.11          &-               \\\hline
\textbf{RoR-3-WRN58-4+SD (13.3M)}         &\textbf{3.77}                 &\textbf{19.73}          &\textbf{1.59}               \\\hline

\end{tabular}
\end{table}
Table~\ref{tab:DFMethods} compares the state-of-the-art methods on CIFAR-10/100 with SVHN, where we achieved overwhelming results. We obtain these results via a simple concept in which the residual mapping of residual mapping was expected to be easier to optimize. No complicated architectures or any other tricks were used. We only added two shortcut levels with thousands of parameters, which made better optimized the original residual networks. For data augmentation, RoR only used naive translation and horizontal flipping, even though other methods often adopted more complicated data augmentation techniques. Our 164-layer RoR-3 had an error of 4.86\% on CIFAR-10, which was better than the 4.92\% of 1001-layer Pre-ResNets with the same batch size. Our 164-layer RoR-3+SD had an error of 22.47\% on CIFAR-100, which was better than the 22.71\% of the 1001-layer Pre-ResNets. Most importantly, it is not only compatible with the original ResNets but also with other kinds of residual networks (Pre-ResNets and WRN). The performance of our Pre-RoR-3 and RoR-3-WRN outperformed original Pre-ResNets and WRN. Particularly, our RoR-3-WRN58-4+SD obtained a single-model error of 3.77\% on CIFAR-10, 19.73\% on CIFAR-100 and 1.59\% on SVHN, which are now state-of-the-art performance standards, to the best of our knowledge. These results demonstrate the effectiveness and versatility of RoR. No matter what kind of basic residual networks are available, RoR can always achieve better results than its basic residual networks with the same number of layers.

\par 
Although some variants of residual networks (WRN and CRMN) or other CNNs (FractalNet) can achieve competitive results, the number of parameters in these models is too large (as shown in Table~\ref{tab:DFMethods}). Through experiments and analysis, we argue that our RoR method can outperform other methods with a similar order of magnitude parameters. Our RoR models with only 3M parameters (Pre-RoR-3-164+SD, RoR-3-WRN40-2+SD and Pre-RoR-3-218+SD) can outperform FractalNet (30M parameters) and CRMN-28 (more than 40M parameters) on CIFAR-10. Our RoR-3-WRN40-4+SD model with 8.9M parameters can outperform all of the exiting methods. Our best RoR-3-WRN58-4+SD model with 13.3M parameters achieved the new state-of-the-art performance. We also contend that a better performance can be achieved by using additional depths and widths.

\subsection{ImageNet Classification}

\begin{table}[!t]
\renewcommand{\arraystretch}{1.3}
\caption{Validation Error (\%, 10-crop testing) on ImageNet by ResNets and RoR-3 with Different Depths}
\label{tab:imagenet}
\centering
\begin{tabular}{|l|c|c|}
\hline
Method                                       &Top-1 Error             &Top-5 Error  \\ \hline\hline
ResNets-18~\cite{gross2016facebookres}       &28.22                                  &9.42               \\\hline
\textbf{RoR-3-18}                            &\textbf{27.84}                                   &\textbf{9.22}                \\\hline
ResNets-34~\cite{he2015resnets}              &24.52                                  &7.46               \\\hline
ResNets-34~\cite{gross2016facebookres}       &24.76                                  &7.35               \\\hline
\textbf{RoR-3-34}                            &\textbf{24.47}                                   &\textbf{7.13}                \\\hline
ResNets-101~\cite{he2015resnets}             &21.75                                  &6.05               \\\hline
ResNets-101~\cite{gross2016facebookres}      &21.08                                  &5.35               \\\hline
\textbf{RoR-3-101}                           &\textbf{20.89}                                   &\textbf{5.24}               \\\hline
ResNets-152~\cite{he2015resnets}             &21.43                                  &5.71               \\\hline
ResNets-152~\cite{gross2016facebookres}      &20.69                                  &5.21               \\\hline
\textbf{RoR-3-152}                           &\textbf{20.55}                                   &\textbf{5.14}                \\\hline
\end{tabular}
\end{table}
The preceding data sets were all small scale and low-resolution image data sets. We also conducted experiments on large scale and high-resolution image data set. We evaluated our RoR method on the ImageNet 2012 classification data set~\cite{Russ2014imagenetchallenge}, which contains 1.28 million high-resolution training images and 50,000 validation images with 1000 object categories. During training of RoR, we noticed that RoR is slower than ResNets. So instead of training  RoR from scratch, we used the ResNets models from~\cite{gross2016facebookres} for pretraining. ResNets for ImageNet required 64, 128, 256, 512 filters (Basic Residual Block) or 256, 512, 1024, 2048 filters (Bottleneck Residual Block) sequentially in the convolutional layers, and each kind of filter had a different number of residual blocks, which formed four residual block groups. We constructed RoR-3 models based on pretrained ResNets models by adding first-level and second-level shortcuts as discussed in Section III. The weights from pretrained ResNets models remained unchanged, but the new added weights were initialized as in~\cite{he2015prelu}. We used SGD with a mini-batch size of 128 (18 layers and 34 layers) or 64 (101 layers) or 48 (152 layers) for 10 epochs to fine-tune RoR. The learning rate started from 0.001 and was divided by a factor of 10 after epoch 5. For data augmentation, we used scale and aspect ratio augmentation~\cite{gross2016facebookres}. Both Top-1 and Top-5 error rates with 10-crop testing were evaluated. In addition, SD was not used here because SD made RoR difficult to converge on ImageNet. From Table~\ref{tab:imagenet}, our implementation of residual networks achieved a state-of-the-art performance compared to ResNets methods for single model evaluation on validation data set. These experiments verified the effectiveness of RoR on large scale and high-resolution image data set.

\subsection{Further Analysis}
In preceding section, the RoR method achieved state-of-the-art results when working with all the best known data sets used for image classification. This final section analyzes the characteristics of the RoR method. Different strategies were adopted for different data sets. On CIFAR-10, RoR alone achieved the best results, as shown in Fig~\ref{fig:RoR-310}. On CIFAR-100 and SVHN, the residual networks encountered severe overfitting. Due to the increasing parameters and extra shortcut levels, using RoR on CIFAR-100 and SVHN proved to increase overfitting which offset the advantages of RoR. However, RoR still improved ResNets marginally. Therefore, we used SD method to reduce overfitting, and both ResNets and RoR benefited from SD. Our RoR+SD outperformed ResNets+SD, as shown in Fig.~\ref{fig:RoR-3100}, Fig.~\ref{fig:WRN100} and Fig.~\ref{fig:SVHN}, as RoR can be sufficiently utilized due to SD. For ImageNet, because of the difficulties to train from scratch, we fine-tuned RoR based on original ResNets models. Because SD slowed the converging process of fine-tuning, so we did not employ SD for ImageNet. By doing so, we achieved better performance than ResNets on ImageNet as well. The final analysis showed good results, which were attributed not only to RoR, but also to the different strategies used in the different data sets. On the other hand, different kinds of residual networks also benefited from RoR, such as Pre-ResNets and WRN. In conclusion, we argue that RoR not only represents a network structure such as ResNets, Pre-ResNets and WRN, but is also an effective complement to the residual-networks family.

\section{Conclusion}
This paper proposes a new Residual networks of Residual networks architecture (RoR), which was proved capable of obtaining a new state-of-the-art performance on CIFAR-10, CIFAR-100, SVHN and ImageNet for image classification. Through empirical studies, this work not only significantly advanced the image classification performance, but can also provided an effective complement to the residual-networks family in the future. In other words, any residual network can be improved by RoR. Hence, RoR has a good prospect of successful application on various image recognition tasks.


%

\appendices

\section*{Acknowledgment}
The authors would like to thank the editor and the anonymous reviewers for their careful reading and valuable remarks.

\ifCLASSOPTIONcaptionsoff
  \newpage
\fi



%

\begin{IEEEbiography}[{\includegraphics[width=1in,height=1.25in,clip,keepaspectratio]{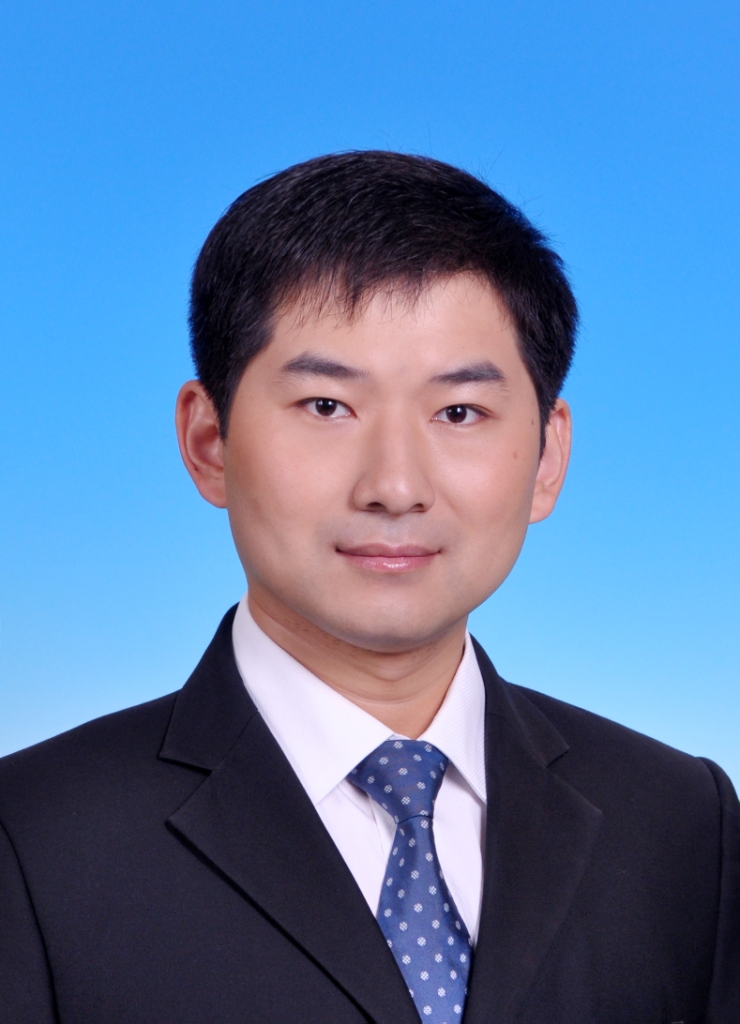}}]{Ke Zhang}
is an associate professor at North China Electric Power University, Baoding, China. He received the M.E. degree in signal and information processing from North China Electric Power University, Baoding, China, in 2006, and the Ph.D degree in signal and information processing from Beijing University of Posts and Telecommunications, Beijing, China, in 2012. His research interests include computer vision, deep learning, machine learning, robot navigation, natural language processing and spatial relation description.
\end{IEEEbiography}

\begin{IEEEbiography}[{\includegraphics[width=1in,height=1.25in,clip,keepaspectratio]{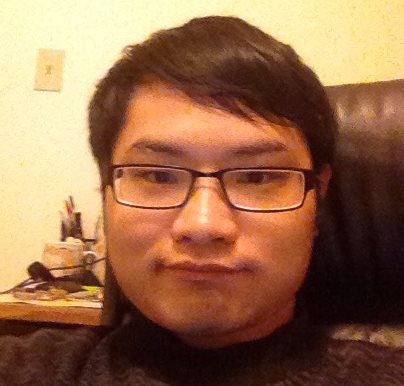}}]{Miao Sun}
received the B.E. degree in automation from the University of Science and Technology of China, in 2011, and the M.S. degree in electrical and computer science from University of Missouri in 2014. He is currently completing the Ph.D degree work at University of Missouri. His research interests include computer vision with special interests in object detection, image classification, and activity analysis, and deep learning with special interests in convolutional networks and hierarchical models.
\end{IEEEbiography}

\begin{IEEEbiography}[{\includegraphics[width=1in,height=1.25in,clip,keepaspectratio]{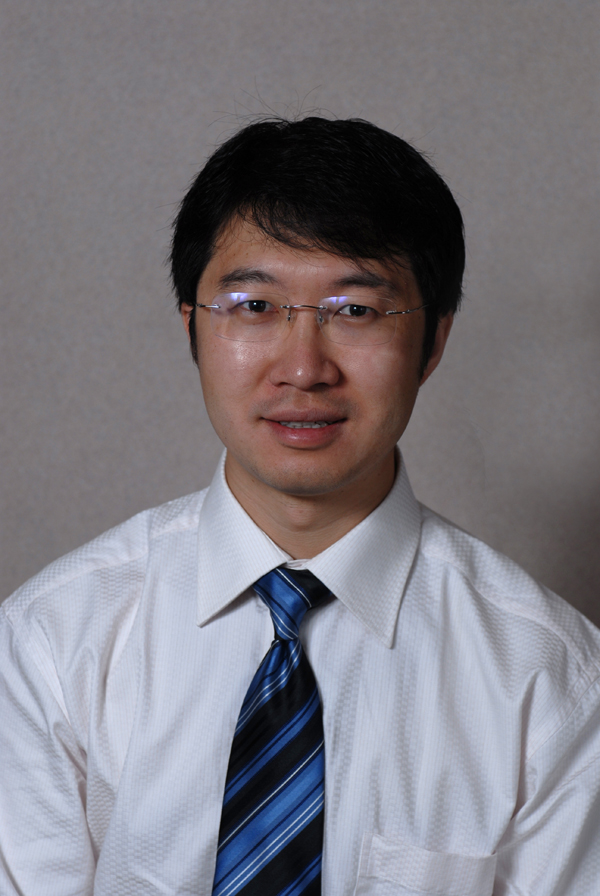}}]{Tony X. Han}
received the B.S. degree with honors in Electrical Engineering Department and Special Gifted Class from Jiaotong University, Beijing, China in 1998, M.S. degree in electrical and computer engineering from the University of Rhode Island, RI, in 2002, and Ph.D degree in electrical and computer engineering from the University of Illinois at Urbana-Champaign, IL, in 2007. He then joined the Department of Electrical and Computer Engineering at the University of Missouri, Columbia, MO, in August 2007. Currently, he is an associate professor of the Department of Electrical and Computer Engineering.
\end{IEEEbiography}

\begin{IEEEbiography}[{\includegraphics[width=1in,height=1.25in,clip,keepaspectratio]{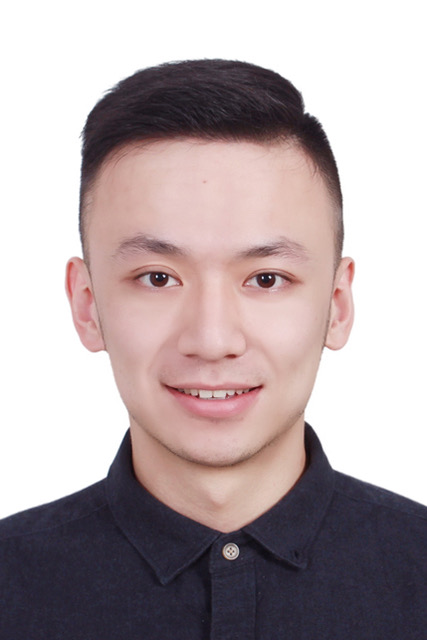}}]{Xingfang Yuan}
received the B.S. degree in electronic science and technology from Zhejiang University, Hangzhou, China in 2015. Currently, he is a Ph.D candidate at the Department of Electrical and Computer Engineering at University of Missouri, Columbia, MO. His research interests include computer vision and deep learning.
\end{IEEEbiography}

\begin{IEEEbiography}[{\includegraphics[width=1in,height=1.25in,clip,keepaspectratio]{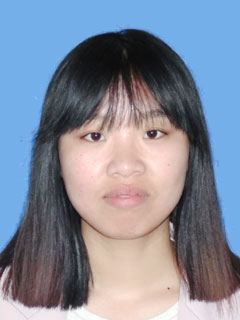}}]{Liru Guo}
received the B.S. degree in telecommunications engineering from Hebei University Of Technology, China, in 2015. Currently, she is a postgraduate student and pursuing for master degree in communication and information engineering from North China Electric Power University. Her research interests include computer vision and deep learning.
\end{IEEEbiography}

\begin{IEEEbiography}[{\includegraphics[width=1in,height=1.25in,clip,keepaspectratio]{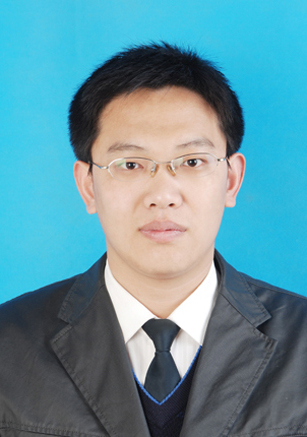}}]{Tao Liu}
is an associate professor at North China Electric Power University, Baoding, China. He received the M.E. degree in optics from Hebei University, Baoding, China, in 2006, and the Ph.D degree in electromagnetic field and microwave technology from Beijing University of Posts and Telecommunications, Beijing, China, in 2009. His research interests include information processing and optical fiber communication.
\end{IEEEbiography}

%






\end{document}